%% file: main.tex
\documentclass[10pt,twocolumn,letterpaper]{article}

\usepackage[pagenumbers]{cvpr} %

\input{preamble}

\usepackage{booktabs}
\usepackage{graphicx}
\usepackage[table,xcdraw]{xcolor}
\usepackage{amsmath} 
\usepackage{amssymb}

\usepackage{multirow}
\usepackage{xfp}
\usepackage{overpic}

\usepackage{arydshln}

\usepackage{tikz}
\usetikzlibrary{spy, calc} %

\definecolor{cvprblue}{rgb}{0.21,0.49,0.74}
\usepackage[pagebackref,breaklinks,colorlinks,allcolors=cvprblue]{hyperref}

\def\paperID{271} %
\def\confName{CVPR}
\def\confYear{2026}

\title{Elastic3D: Controllable Stereo Video Conversion with Guided Latent Decoding}

\author{
    Nando Metzger$^{1,2,}$* \quad
    Prune Truong$^2$ \quad
    Goutam Bhat$^2$ \quad
    Konrad Schindler$^1$ \quad
    Federico Tombari$^{2,3}$ \\[2mm]
    $^1$ETH Zurich \qquad 
    $^2$Google \qquad 
    $^3$TU Munich \\
}

\begin{document}
\input{src/figs/00_teaser}

{
  \renewcommand{\thefootnote}{\fnsymbol{footnote}} %
  \footnotetext[1]{Work conducted during an internship at Google.}
}

\input{sec/0_abstract}

\input{sec/1_intro}

\input{sec/2_related_work}

\input{sec/3b_method}

\input{sec/3a_preliminary}
\input{sec/4_evaluation_protocol}
\input{sec/6_results}

\input{sec/7_conclusion}

{
    \small
    \bibliographystyle{ieeenat_fullname}
    \bibliography{main}
}

\input{sec/X_suppl}

\end{document}

%% file: preamble.tex
\definecolor{almond}{rgb}{0.94, 0.87, 0.8}

\newcommand{\parsection}[1]{\vspace{1mm}\noindent\textbf{#1:}~}

\newcommand{\ours}{Elastic3D\xspace}

%% file: src/figs/00_teaser.tex
\twocolumn[{%
    \renewcommand\twocolumn[1][]{#1}%
    \maketitle
    \begin{center}
      \vspace{-0.9em}
      \includegraphics[width=\textwidth]{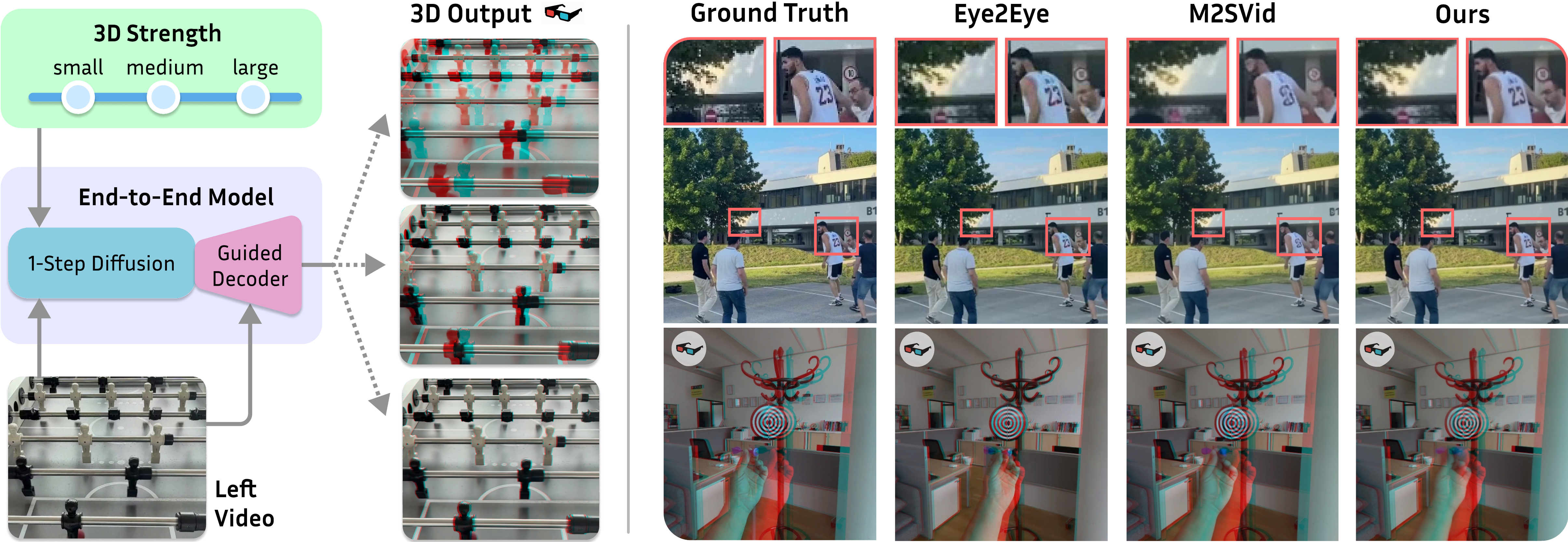}
      \captionsetup{type=figure}
      \vspace{-1.5em}
      \captionof{figure}{
        We introduce a direct, warping-free and feed-forward approach for mono-to-stereo video conversion. Our latent diffusion model (left) bypasses the need for depth estimation, generating the right views directly from the input left views in a single step. We achieve control over the 3D strength by conditioning on a scalar disparity factor, and employ a left-view guided VAE decoder to preserve high-frequency details and eliminate binocular rivalry artifacts. As opposed to M2SVid~\cite{m2svid}, our approach seamlessly transfers high-frequency details from the left to the right videos (top-right). Unlike Eye2Eye~\cite{eye2eye}  which lacks the ability to control the 3D strength in generated content and limits its practical use, we can generate videos for any disparity distribution, as shown in the anaglyph (bottom-right).
      }\label{fig:teaser}
      \vspace{0.3em}
     \end{center}%
}]

%% file: sec/0_abstract.tex
\begin{abstract}
The growing demand for immersive 3D content calls for automated monocular-to-stereo video conversion. We present a controllable, direct end-to-end method for upgrading a conventional video to a binocular one. Our approach, based on (conditional) latent diffusion, avoids artifacts due to explicit depth estimation and warping. The key to its high-quality stereo video output is a novel, guided VAE decoder that ensures sharp and epipolar-consistent stereo video output. Moreover, our method gives the user control over the strength of the stereo effect (more precisely, the disparity range) at inference time, via an intuitive, scalar tuning knob.
Experiments on three different datasets of real-world stereo videos show that our method outperforms both traditional warping-based and recent warping-free baselines and sets a new standard for reliable, controllable stereo video conversion.
Please check the project page for the video samples \url{elastic3d.github.io}.
\end{abstract}

%% file: sec/1_intro.tex
\section{Introduction}
\label{sec:intro}

The proliferation of virtual and augmented reality hardware has created a surge in demand for immersive 3D content.
With the vast majority of video content being monocular, computational mono-to-stereo conversion has become an important capability.
The dominant paradigm follows a two-step warp-and-refine process: first estimate scene depth from the monocular input; second, reproject the input to a new viewpoint via the depth map~\cite{m2svid,stereocrafter,svg, restereo}.
This warping process inevitably creates holes and artifacts, particularly in disoccluded regions (areas newly visible in the target view), which must then be inpainted or refined to produce a complete image.

This approach, however, faces challenges. The reliance on a separate depth estimator makes the pipeline more complex and brittle. 
Its quality often critically depends on the accuracy of the intermediate (monocular) depth estimator, which can struggle with certain scenarios, e.g. thin structures and non-lambertian surfaces.
Moreover, most of these approaches~\cite{stereocrafter, m2svid} rely on Latent Diffusion Models (LDMs), operating in a compressed latent space for efficiency.
However, their general-purpose decoders create an information bottleneck, failing to reconstruct details from the source view (see the missing and distorted details of M2SVid in Fig.~\ref{fig:teaser} right). This loss of detail, or the hallucination of new, inconsistent details, leads to a disorienting visual artifact known as \textit{binocular rivalry}~\cite{binocular_rivalry}, where the viewer's eyes are presented with conflicting information.

To overcome these issues, the recent Eye2Eye~\cite{eye2eye} method directly generates the second viewpoint conditioned on the monocular input, sidestepping explicit depth estimation and warping.
While promising, it has left a major challenge unsolved, namely, the lack of geometric control. A key advantage of warp-based methods is that users can easily adjust the 3D effect by simply scaling the intermediate depth map. In contrast, Eye2Eye is trained for a fixed, implicit baseline and offers no mechanism to control the stereo disparity, limiting its artistic and practical flexibility. Moreover, while its reliance on a two-stage refinement approach with a pixel-space diffusion model largely improves the stereoscopic fidelity of the generated video, it is prohibitively slow. 

Our primary contribution is a surprisingly straightforward, warping-free model for stereo video conversion that elegantly solves both of these challenges. 
The model, based on Stable Video Diffusion~\cite{blattmann2023stable}, directly synthesizes the second view in a latent space and features two innovations:
(1)~a conditioning mechanism that provides intuitive user control over the strength of the stereo effect by simply setting a scalar (continuous) ``median disparity'' at inference time. And, 
(2)~a new guided VAE decoder, which receives high-resolution information from the source view, bypassing the latent bottleneck, to support the reconstruction of fine details and minimize binocular rivalry.

The field of video conversion lacks a comprehensive evaluation protocol for the entire mono-to-stereo conversion pipeline. Previous approaches focus only on some aspects, such as inpainting quality. 
Hence, as a necessary side effect of our approach, we contribute a comprehensive evaluation protocol for the entire stereo conversion pipeline.
Our framework is designed to achieve a holistic quality assessment, by quantifying and comparing stereo videos along four essential dimensions: overall quality taking into account 3D-strength controllability, stereoscopic fidelity, geometric correctness, and temporal consistency.

To summarize, the contributions of this paper are:
(1) a direct, \emph{warping-free and feed-forward model for stereo video conversion} that does not explicitly depend on any depth estimation;
(2) a novel \emph{conditioning method that gives the user control over the stereoscopic disparity}, trained only on the wild rectified video data without known calibration;
(3) a \emph{guided VAE architecture that minimizes binocular rivalry} in the latent diffusion framework;
(4) an improved, \emph{comprehensive evaluation protocol} that covers the geometric, perceptual and temporal dimensions of stereo video quality.
Our approach to stereo video conversion sets a new state of the art across three in-the-wild datasets.

%% file: sec/2_related_work.tex
\section{Related Work}
\label{sec:related}

We situate our work at the intersection of stereo image synthesis, controllable latent diffusion, and guided decoding.

\subsection{Stereo Conversion}

Classic conversion approaches typically follow a ``depth-then-warp" paradigm via depth-image-based rendering (DIBR)~\cite{konrad20122d, zhang20113d, fehn2004depth}. While the depth estimation stage has been significantly improved by modern estimators~\cite{marigold,midas,dpt,depthcrafter, depthanything,ke2023repurposing,unidepth,unidepthv2, depthpro}, subsequent warping steps still introduce artifacts such as disocclusions and artifacts on non-Lambertian surfaces. Even recent diffusion-based methods often retain explicit warping, inheriting these limitations~\cite{wang2024stereodiffusion, genstereo}. Similar trends are observed in video generation~\cite{zhang20113d, SplatDiff, m2svid, stereocrafter, svg, restereo}. Building on pioneering warping-free works like Deep3D~\cite{deep3d} and Eye2Eye~\cite{eye2eye}, our method avoids warping artifacts by directly synthesizing the target view in latent space, advancing the state of the art in robustness, controllability, and detail preservation.

\subsection{Diffusion Models}

Denoising diffusion probabilistic models (DDPMs)~\cite{ho2020denoising, song2021denoising} have become the state-of-the-art in generative modeling. Latent Diffusion Models (LDMs)~\cite{rombach2022high} scale the capabilities of DDPMs~\cite{ho2020denoising, song2021denoising} by operating in a compressed latent space, enabling efficient high-resolution image~\cite{podell2023sdxl, saharia2022photorealistic} and video synthesis~\cite{blattmann2023stable, ho2022video}. LDMs consist of two main components: an autoencoder (typically a VAE~\cite{kingma2013auto}) and a diffusion model trained in the autoencoder's latent space. The VAE's encoder compresses a high-resolution image $x$ into a lower-dimensional, usually quantized latent representation $z$, and its decoder reconstructs the image $\hat{x}$ from $z$. The diffusion model learns to denoise this latent $z$. While this is highly efficient, the compression to $z$ creates an information bottleneck, discarding information details that the decoder must regenerate.
Our method leverages the efficiency, temporal pretraining and generative priors of video LDMs while introducing a guided decoding mechanism to explicitly bypass this bottleneck and recover lost detail.

\subsection{Guided Decoding}

Re-injecting high-fidelity source information is standard practice in guided super-resolution~\cite{hui2016depth, lutio2019guided, metzger2023guided} and pansharpening~\cite{vivone2014critical, chavez1991comparison}. In diffusion models, this is achieved via convolutional skip connections for image translation~\cite{parmar2024one, yang2025any} or ``Texture Bridges'' for view synthesis~\cite{SplatDiff}. Geometric constraints also serve as effective guidance, notably in Epipolar Transformers~\cite{shin20193d, epipolartransformer}. Most relevant to our approach, \cite{wu2025direct} utilizes epipolar attention for simultaneous multi-view generation. We distinguish our work by integrating a lightweight epipolar attention module directly into the VAE decoder; this creates structured skip connections that bypass the bottleneck by sampling details along geometrically plausible lines.

\subsection{Controllable Generation}
Significant progress has been made in controlling generative models using spatial conditions, such as depth or edge maps in ControlNet~\cite{zhang2023adding} and T2I-Adapters~\cite{mou2024t2i}. Alternatively, cross-attention mechanisms allow for conditioning on learned embeddings to control style and object identity~\cite{gal2022image, ruiz2023dreambooth, ye2023ip} or geometric parameters like camera trajectories~\cite{yu2025trajectorycrafter}. However, these methods do not address stereoscopic layouts. Our work is the first to propose a mechanism for controlling the stereo effect via a simple and intuitive scalar disparity proxy.

%% file: sec/3b_method.tex
\section{What Makes a Good Stereo Model?}
\label{sec:what_makes_a_good_model}

Given an input ``left-eye'' video $V_L \in \mathbb{R}^{N \times H \times W \times 3}$, consisting of $N$ frames of resolution $H \times W$, the goal of monocular-to-stereo video conversion is to synthesize a ``right-eye'' video $\hat{V}_R \in \mathbb{R}^{N \times H \times W \times 3}$ that shows the same scene as if viewed from a horizontally shifted camera position. In this section, we first describe the desired properties of such a stereo conversion model.
For a comfortable and immersive 3D experience, the generated video pair $(V_L, \hat{V}_R)$ must satisfy the following key properties:

\begin{enumerate}

    \item \textbf{Geometric Correctness.} The stereo pair must adhere to the epipolar constraint and provide plausible depth perception corresponding to a rectified stereo set-up~\cite{zhou2025perceptual,lambooij2011visual}, with accurate relative depth ordering.

    \item \textbf{3D-Effect Control.} A stereo conversion model must provide continuous (ideally intuitive) control over the 3D strength. This is crucial for both content creators for creative control, and end-users to adjust the immersion level for personal comfort~\cite{terzic2016methods}.
    
    \item \textbf{Stereoscopic Fidelity and Detail Preservation.} To avoid distracting binocular rivalry~\cite{binocular_rivalry}, each synthesized view must losslessly transfer textures from the source left view~\cite{blake2002visual,shibata2011zone,shao2013perceptual}.
    Any inconsistency in shared regions degrades the user experience~\cite{zhou2025perceptual}.
    
    \item \textbf{Plausible Disocclusion Handling.} The model must realistically inpaint regions that were occluded in the input left view by utilizing information from the neighboring frames whenever possible~\cite{watson2020learning,Shih3DP20,mehl2024stereo,shi2024immersepro}.

    \item \textbf{Temporal Stability.} The generated video must be temporally consistent for a comfortable viewing experience~\cite{terzic2016methods,lambooij2011visual}.
\end{enumerate}

Achieving all five properties simultaneously is the central challenge. Most existing stereo conversion methods utilize pretrained video diffusion models~\cite{stereocrafter,svg,m2svid,eye2eye} which inherently provide strong inpainting capability and temporal stability. The warp-based methods~\cite{svg,stereocrafter, m2svid} rely on explicit depth estimation to reproject the input left views, providing effective `3D-Effect Control'. However, this requires using a separate monocular depth model, increasing system complexity and latency. Moreover, the geometric correctness of the method is bounded to that of the depth model.  These methods also suffer from binocular rivalry due to the lossy compression introduced by latent diffusion models. The recent Eye2Eye~\cite{eye2eye} approach mitigates the dependency on a separate monodepth model by directly synthesizing the right view. However, it does not provide any mechanism to control the 3D effect. Furthermore, it is prohibitively slow due to a two-stage refinement process and the usage of a multi-step pixel-space diffusion model~\cite{bar2024lumiere}. In this work we aim to address the aforementioned issues.

\begin{figure} [t]
    \centering
    \includegraphics[width=\linewidth]{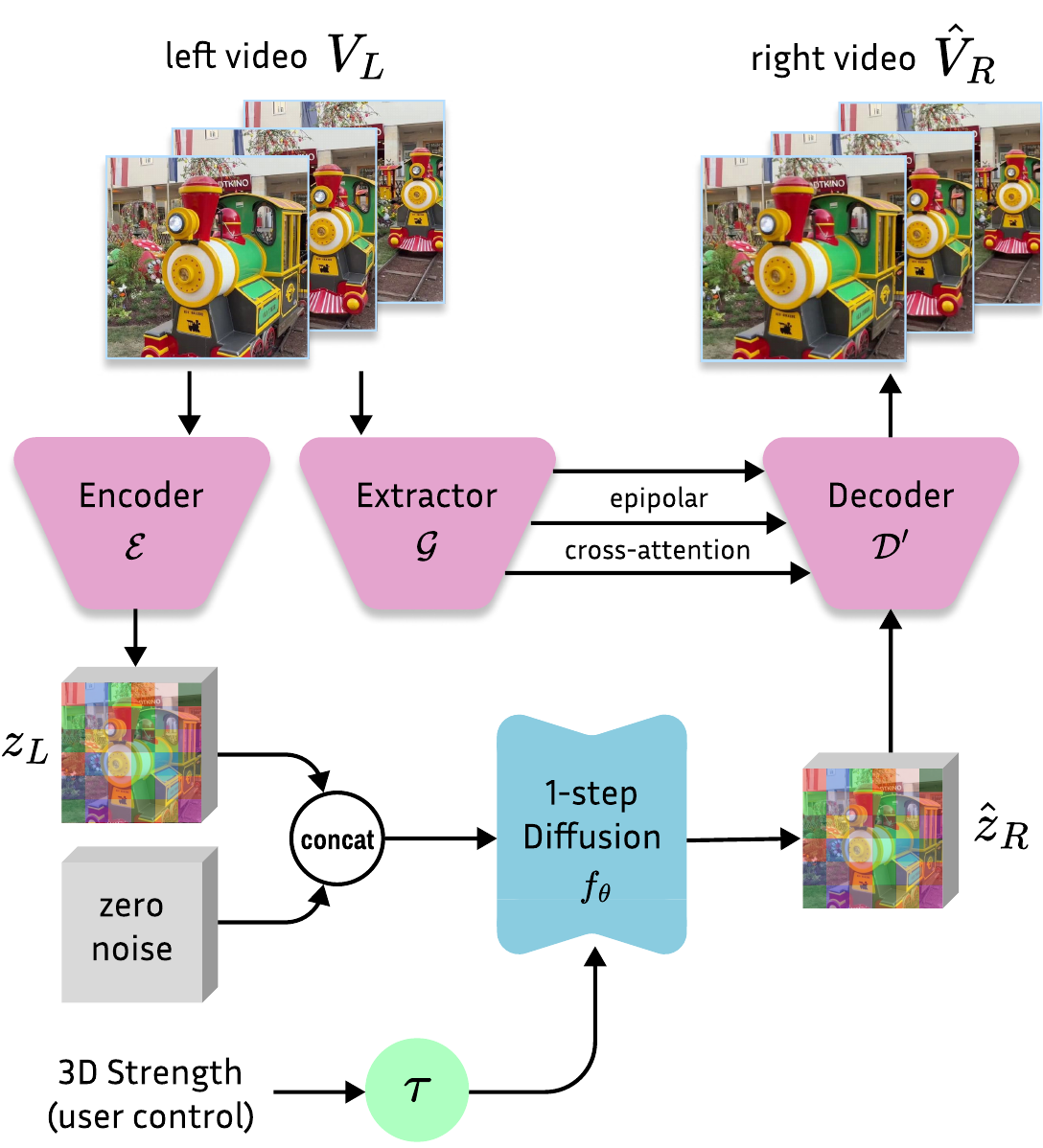}
    \caption{\textbf{Inference Pipeline.}
    A frozen VAE Encoder $\mathcal{E}$ computes the latent code $z_L$ of the input video $V_L$. The synthesis network $f_\theta$ then generates the right-view latent $\hat{z}_R$, conditioned on $z_L$ and on a 3D strength control token $\tau(\delta)$ (Sec.~\ref{ssec:controllable_warping_free}). 
    Finally, our Guided Decoder $\mathcal{D}'$ (Sec.~\ref{ssec:guided_decoding}) renders the high-fidelity output $\hat{I}_R$, using both the generated latent $\hat{z}_R$ and the original video $V_L$ as guidance. 
    }
    \label{fig:method_overview}
    \vspace{-4mm}
\end{figure}

\section{Methodology}
\label{sec:method}

We present a novel, warping-free latent diffusion pipeline that aims to satisfy the key stereo model requirements listed in Sec.~\ref{sec:what_makes_a_good_model}. An overview of our inference pipeline is shown in Fig.~\ref{fig:method_overview}. Given an input video, our 1-step latent diffusion model directly generates the right camera views, without relying on any explicit depth input (Sec.~\ref{ssec:feed-forward}). Crucially, our diffusion model is conditioned on a scalar disparity factor, enabling seamless control over the amount of `3D Effect' in the generated stereo video (Sec.~\ref{ssec:controllable_warping_free}). Furthermore, we introduce a left-view guided VAE decoder %
to effectively transfer the high-frequency details from the input left view to the generated right view, mitigating binocular rivalry  (Sec.~\ref{ssec:guided_decoding}). 
We describe these key contributions in the next sections. 

\subsection{Feed-forward Warping-Free Synthesis}
\label{ssec:feed-forward}

We build our stereo conversion model upon a latent diffusion model, Stable Video Diffusion~\cite{blattmann2023stable}. We utilize the 1-step denoising process~\cite{marigolde2e}, effectively turning the latent denoiser into a feed-forward model.
Unlike warp-based pipelines~\cite{m2svid, stereocrafter} which often use the diffusion model to \textit{refine} a warped video, our model $f_\theta$ \textit{directly} synthesizes the entire right-video latent $\hat{z}_R$ conditioned on the left-video latent $z_L$ and a disparity-conditioning signal $\tau(\delta)$ (Sec.~\ref{ssec:controllable_warping_free}).
This yields an output:
\begin{equation}
    \hat{V}_R = \mathcal{D'}(f_\theta(\mathbf{0}, z_L, \tau(\delta)), V_L), \quad  \text{with} \quad  z_L=\mathcal{E}(V_L)
\end{equation}

\noindent where $\mathcal{E}$ is the VAE encoder and $\mathbf{0}$ is a zero vector. $\tau(\delta)$ refers to our \emph{Disparity Conditioning Mechanism} (Sec.~\ref{ssec:controllable_warping_free}) while $\mathcal{D'}$ is the \emph{Guided Decoder}, described in Sec.~\ref{ssec:guided_decoding}. Our feed-forward approach can efficiently generate the right views without performing numerous denoising iterations. It also enables training with image-space losses, improving the quality of the generated views, as shown in~\cite{m2svid}.

\subsection{3D-Controllable Synthesis}
\label{ssec:controllable_warping_free}

As discussed in Sec.~\ref{sec:what_makes_a_good_model}, it is important that a stereo conversion model provides an intuitive control over the strength of the 3D effect in the generated videos.
We introduce a novel disparity conditioning mechanism to enable this in our warp-free method.
Our synthesis model $f_\theta$ is conditioned on a scalar disparity proxy, $\delta \in \mathbb{R}$, which controls the amount of pixel disparity between the input view and the generated right view. This provides a simple and intuitive `3D strength' knob at inference and, crucially, removes any dependency on camera parameters, enabling training on uncalibrated `in-the-wild' rectified video pairs $(V_L, V_R)$.

We project $\delta$ into a token embedding $\tau(\delta)$ and inject it into the model's spatial attention layers:

\begin{equation}\label{eq:conditioning}
    \hat{z}_R = f_\theta(\mathbf{0}, z_L, \tau(\delta))
\end{equation}

\parsection{Conditioning $\delta$ during training} We compute $\delta$ using the ground-truth disparity maps $D_{L \to R}^0 \in \mathbb{R}^{H \times W}$ from the first frame of the left to the right video.
\begin{equation}\label{eq:conditioning-def}
    \delta = \text{P}_{50}(D_{L \to R}^0)
\end{equation}
The median disparity is chosen as it is insensitive to outliers and provides an interpretable measure of the scene's overall stereo effect. However, we found that both the average and the max disparity are also adequate choices (see Appendix).

\input{src/figs/delta_conditioning}

\parsection{Conditioning $\delta$ at inference}  A user can control the 3D effect by adjusting the $\delta$ in pixels, as shown in Fig.~\ref{fig:delta-conditioning}.

\parsection{Training strategy} To train the model, we use pairs of left $V_L$ and ground-truth right $V_R$ frames. The model is trained to minimize a composite loss function for 1-step diffusion models~\cite{marigolde2e,m2svid} consisting of equally weighted L2-latent loss, and pixel space L1, SSIM, and LPIPS objectives (backpropagated through the frozen VAE decoder).

To teach the model to generalize beyond the narrow range of baselines found in typical datasets, we employ a data augmentation strategy.
For each sample, we randomly scale its ground-truth disparity map $D_{L \to R}^0$ by a factor $s$, generate a new warped view as the right-view GT using a simple forward warp, and use the corresponding scaled disparity map to get the new conditioning token $\tau(s \cdot \delta)$.
We then train on a mixture of original (real) and augmented (synthetic) samples.
For the synthetic pairs, we apply a simple L1 pixel loss, masked to exclude invalid pixels from the forward warp, while real pairs use our full composite loss.

\subsection{Guided Latent Decoding}
\label{ssec:guided_decoding}

We now solve the primary limitation of latent synthesis models: Detail Preservation (Property 2, see Sec.~\ref{sec:what_makes_a_good_model}). The VAE's information bottleneck is the main cause of binocular rivalry. As shown in Fig.~\ref{fig:vae_compression}, the Stable Video Diffusion Decoder struggles to even decode the GT latents, omitting or even hallucinating micro-details. This is unsurprising given the SVD bottleneck's high compression ratio of 1:48.
We introduce a novel guided decoder, $\mathcal{D}'$, which bypasses this bottleneck by drawing details directly from the input left video, as illustrated in the top part of Fig.~\ref{fig:method_overview}.

\parsection{Epipolar-guided decoder} We reformulate decoding as a guided latent upsampling task.
The decoder $\mathcal{D}'$ is conditioned on both the synthesized latent code $\hat{z}_R$ and the original left video $V_L$. An extractor network $\mathcal{G}$, initialized from the VAE encoder, processes $V_L$ to produce frame-wise pyramids of multi-scale guidance feature maps, $\{g_1, ..., g_N\}$, which act as an information reservoir. 

Inspired by novel view synthesis literature~\cite{huang2024epidiff, epipolartransformer}, we introduce a lightweight epipolar attention mechanism. At each upsampling block $i$, for each feature vector $h_i(p)$ in the decoder (query), its cross-attention is computed only over the keys and values from the corresponding epipolar line in the guidance map $g_i$. In our rectified case, this simplifies to a one-dimensional key-query correspondence along the horizontal row.
This geometrically constrained attention $\mathcal{A}_{\text{epipolar}}$ allows the decoder to ``look up" and re-inject the information needed to reconstruct the image.
More explicitly, the refined feature is computed residually:
$$
h_i'(p) = h_i(p) + \mathcal{A}_{\text{epipolar}}(h_i(p), g_i)
$$
\parsection{Light-weight epipolar-attention} The epipolar constraint is not only geometrically motivated but also crucial for efficiency, reducing the complexity of full-cross-attention from a prohibitive $\mathcal{O}(H^2W^2)$ to a more manageable $\mathcal{O}(HW^2)$. For our experiments where the decoder's highest feature map is 512×512, this reduces the memory required for the 16-bit attention matrix from an infeasible 128 GB to just 256 MB.

\input{src/figs/01_vae_compression}

\parsection{Implementation details} We initialize our decoder $\mathcal{D}'$ with the weights of the standard decoder  $\mathcal{D}$. The guidance network $\mathcal{G}$ is initialized with the weights of the encoder $\mathcal{E}$. The output projection of the attention layers $\mathcal{A}$ is initialized with zero weights, so to preserve the mapping of the original decoder at the beginning of the training.

\parsection{Training strategy}
We train the decoder as a separate module to disentangle the task of geometric synthesis from low-level texture reconstruction. This keeps the method modular and plug-and-play with respect to 3rd party synthesis cores $f_\theta$, as shown in Tab.~\ref{tab:ablation_study_model_stereo4d}.
The objective is to reconstruct the ground-truth right view $V_R$ from its latent $z_R = \mathcal{E}(V_R)$, using $V_L$ as guidance.
We use a standard reconstruction loss combining L1 and LPIPS as in~\cite{SplatDiff}.

%% file: src/figs/delta_conditioning.tex
\begin{figure} [t]
    \centering
    
    \includegraphics[width=\linewidth]{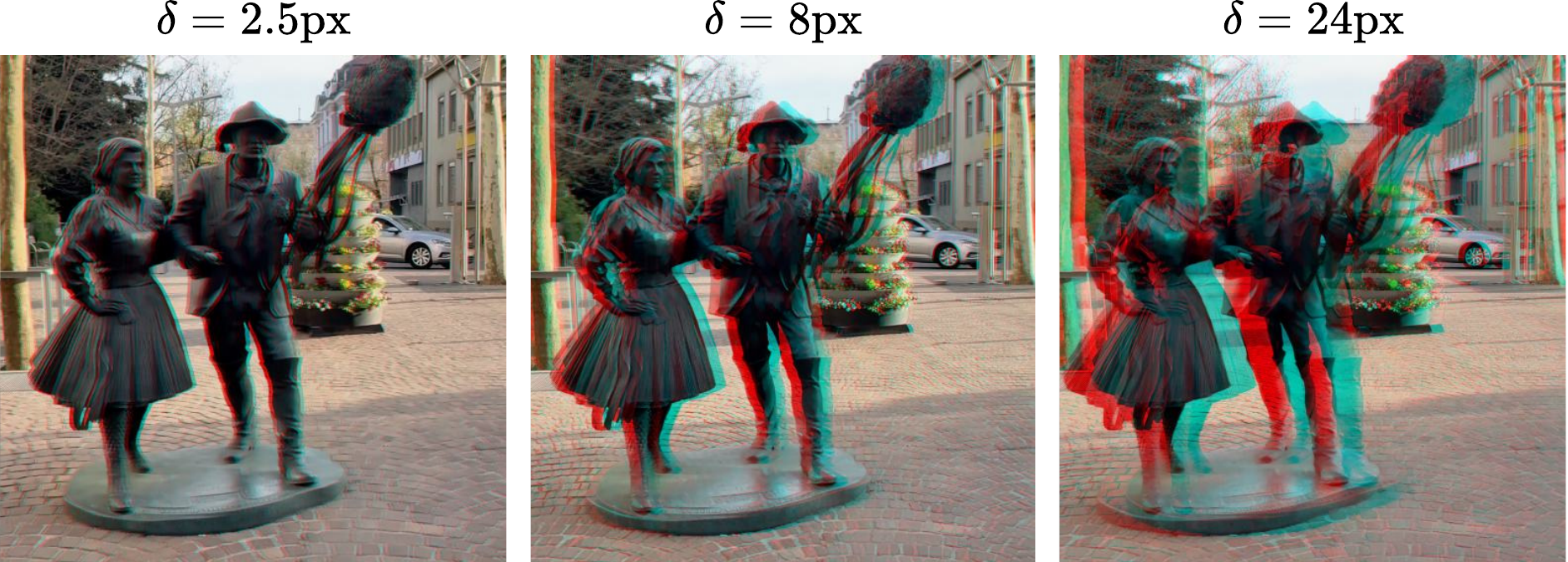}
    \vspace{-7mm}
    \caption{The strength of the stereo effect can be controlled by varying the parameter $\delta$ that acts as a conditioning for the median disparity of the generated video (see Sec.~\ref{ssec:controllable_warping_free}). }
    \label{fig:delta-conditioning}
    \vspace{-4mm}
\end{figure}

%% file: src/figs/01_vae_compression.tex
\begin{figure}[t]
    \centering
    \includegraphics[width=\linewidth]{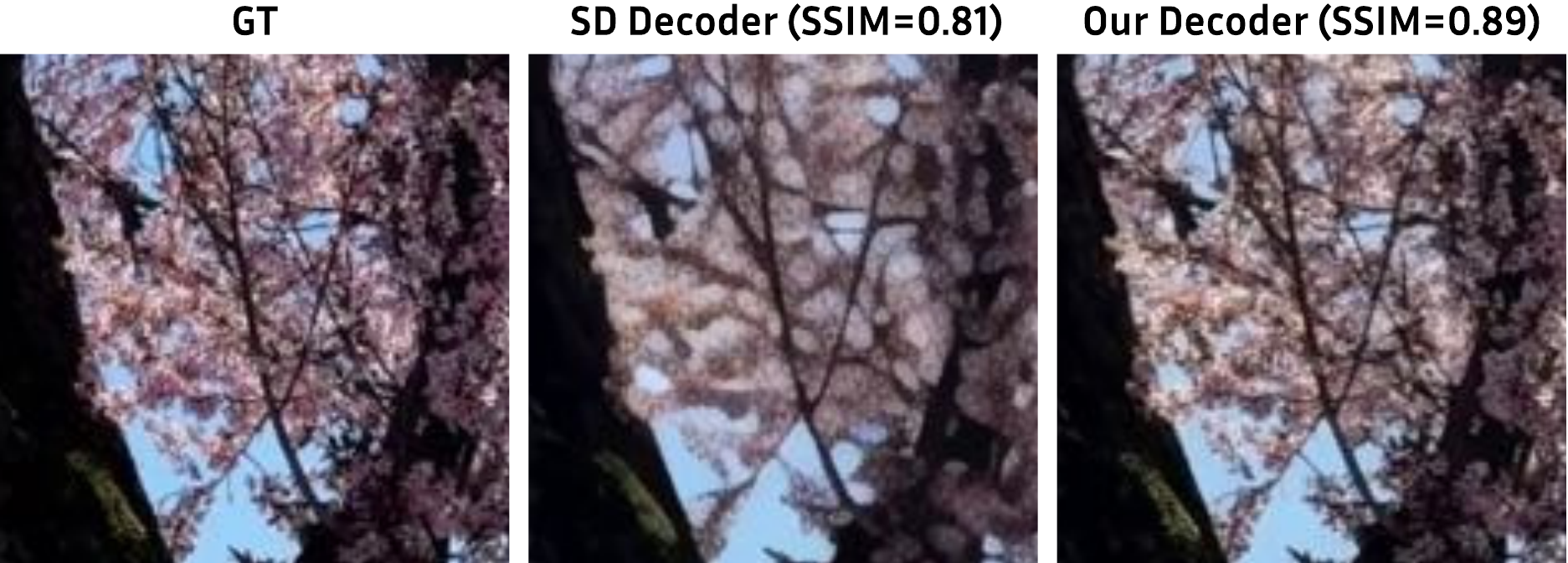}
    \vspace{-7mm}
    \caption{Decoding the ground truth latent. The compression with the standard VAE of an LDM is unsuitable for discriminative tasks.}
    \label{fig:vae_compression}
    \vspace{-5mm}
\end{figure}

%% file: sec/4_evaluation_protocol.tex
\section{Evaluation of Stereo Conversion}
\label{sec:evaluation}

A model is only as good as the benchmark used to measure it. However, there is a lack of a common evaluation protocol for the full stereo conversion task. Recent warp-based methods~\cite{m2svid,genstereo} compute standard image quality metrics between the ground truth and generated right views. However, they utilize stereo depth from the ground-truth image pair~\cite{m2svid,genstereo} to warp the source images. This leaks target-view information, fundamentally changing the task from robust geometric inference to simple guided inpainting. Other methods~\cite{svg,eye2eye,m2svid} utilize human user studies, which are very hard to scale. These methods also do not evaluate the 3D controllability of the stereo generation, which is a key requirement as described in Sec.~\ref{sec:what_makes_a_good_model}.

To address this, we propose a standardized ``black-box'' evaluation protocol for the full stereo conversion pipeline. We require all methods -- whether warp-based or end-to-end -- to operate solely on the monocular input $I_L$ at inference time.
For warp-based methods, this means using their intended monocular depth estimator. In order to account for the differences in `3D strength' in the test videos (due to \eg differences in stereo cameras), the methods are also provided access to global 3D information. For the warp-based methods, this entails providing the scale and shift factors to align the relative monocular depths to the pseudo ground truth metric depth. For our approach, we provide the median disparity between left and right views. We evaluate the methods on diverse videos captured from different stereo cameras in order to validate that the approaches can provide effective 3D control.

\subsection{Evaluation Metrics}
We assess this full pipeline using four complementary metric categories.

\parsection{Overall Quality}
To evaluate the overall quality of the left-to-right stereo conversion, we use standard full-reference metrics: Peak Signal-to-Noise Ratio (\textit{PSNR}), Structural Similarity (\textit{SSIM})~\cite{ssim}, and Learned Perceptual Image Patch Similarity (\textit{LPIPS})~\cite{lpips}. Note that these metrics are highly sensitive to pixel-wise alignment between the generated and GT views, and hence evaluate the combination of all properties listed in Sec.~\ref{sec:what_makes_a_good_model}, i.e. geometric correctness, 3D-strength controllability, perceptual fidelity in shared and inpainted regions and temporal stability.

In order to better understand the trade-off of the different methods, we also propose ``component-wise'' metrics, which evaluate only one of the properties listed in Sec.~\ref{sec:what_makes_a_good_model}, while being less sensitive to the others.

\parsection{Stereoscopic Fidelity}
Standard image quality metrics such as PSNR cannot capture specific stereo artifacts such as binocular rivalry, which is caused by mismatches in the left and right images. 
We thus introduce specialized metrics to evaluate this.
\begin{itemize}[leftmargin=*]
    \item \textit{Matchability Error} ($\mathcal{E}_{\text{Match}}$) As a proxy for binocular rivalry, we measure the inconsistency of feature keypoints between views. Using a robust matcher (DeDoDe v2~\cite{dedodev2}), we identify keypoints in $I_L$ that have epipolar-consistent matches in the ground truth ($M_{gt}$) and in the prediction ($M_{pred}$). We define the error as the complement of their Jaccard index,
    \begin{equation}
        \mathcal{E}_{\text{Match}} = 1 - \frac{|M_{gt} \cap M_{pred}|}{|M_{gt} \cup M_{pred}|}
    \end{equation}
    A lower error indicates that the synthesized view maintains consistent, matchable texture details along the correct epipolar geometry, minimizing rivalry.

    \item \textit{Patch-wise PSNR (P-PSNR)} To assess local photometric consistency between input left and generated right, while being robust to small local and global disparity shifts, we measure PSNR between $16\times16$ patches in $I_L$ and their best-matching counterparts found along the horizontal epipolar line in the generated $\hat{I}_R$.
\end{itemize}
Note that as opposed to the earlier perceptual metrics, these metrics are less sensitive to precise pixel-wise alignment between the generated and GT right view and are thus partly decorrelated from the geometric correctness and 3D controllability. Nevertheless, both proposed metrics perform searches strictly along horizontal epipolar lines, inherently penalizing any vertical misalignment or rectification errors.

\parsection{Geometric Correctness} We also validate that the model generates stereo videos with correct geometry and depth ordering. We estimate the dense disparity maps between the generated ($D_{pred}$) stereo pairs using FoundationStereo~\cite{foundationstereo}, and compare it with the (pseudo) ground truth disparity maps.
In order to disentangle the depth ordering errors from incorrect `3D strength' and account for global scale differences, we perform a least-squares alignment between predicted and ground truth disparities and report the Mean Absolute Error between aligned disparities, as \textit{Disp.~err}.

\parsection{Temporal Stability}
For video, consistency over time is paramount. We measure this by computing optical flow fields for both the ground-truth and generated right-view videos using RAFT~\cite{raft}. We report the End-Point Error, measuring the deviation of the generated motion from the true scene motion, which we denote as \textit{Temp.~err}.

%% file: sec/6_results.tex
\section{Experiments}
\label{sec:res}

In this section, we evaluate the quality of the stereoscopic
videos generated by our method, both qualitatively as well
as quantitatively. Further results, analysis, visualizations
and implementation details are provided in the Appendix.

\subsection{Implementation Details}

\parsection{Training data} Following~\cite{m2svid}, we train on the \textbf{Stereo4d} and \textbf{Ego4d} datasets containing stereoscopic videos. Stereo4d comprises diverse real-world scenes and dynamic objects captured by internet videos with a fixed baseline of 63mm, while Ego4d features egocentric footage characterized by large disparities. For our work, we utilize the splits and stereo-rectified version of the data provided by~\cite{m2svid}, at a resolution of $512 \times 512$, with $N=16$. We use the FoundationStereo~\cite{foundationstereo} model to estimate the disparity maps used to compute the 3D strength $\delta$.

\parsection{Evaluation set-up} For in-domain evaluation, we rely on the test-split of Stereo4D. Results on Ego4D are provided in the Supplementary. For out-of-distribution evaluation, we use the Spatial Video Dataset~\cite{SpatialVD2025} containing stereoscopic videos captured from Apple Vision Pro (AVP) and iPhone. The AVP portion features data taken with a similar baseline (63.8mm, i.e.\ close to inter-eye distance) as the training data, while the iPhone data is captured with a significantly lower baseline of 19.2mm.
For the warping-based approaches~\cite{svg, stereocrafter, m2svid}, we extract relative depths with DepthCrafter~\cite{depthcrafter} and align it to the GT disparity, as described in~\ref{sec:evaluation}. The aligned mono-depth is then used to generate the warped right view. For our method, the GT disparity is used to compute the conditioning factor $\delta$ with eq.~\eqref{eq:conditioning-def}.

\subsection{Model Ablations and Analysis}

We first ablate the key components of our approach. More analysis is provided in Appendix.

\input{src/tabs/ablation_conditioning}

\parsection{Warping-free conditioning} In Tab.~\ref{tab:iphone_ablation}, we evaluate the impact of the proposed disparity conditioning (Sec.~\ref{ssec:controllable_warping_free}), which enables controlling the 3D strength in the generated stereoscopic videos.
We train a baseline model without the disparity conditioning on the Stereo4D and Ego4D datasets, both of which contain videos predominantly captured with stereo cameras with baseline close to 63mm.
On the test split of the Stereo4D dataset, which includes videos with similar 3D strength as those in the train split, the baseline without conditioning obtains a PSNR of $25.1$ (Tab.~\ref{tab:iphone_ablation}, top). Adding the conditioning nevertheless allows to get more fine-grained spatial alignment of the generated and GT right videos, leading to better overall metrics ($+1$ dB in PSNR). 
On the Spatial Video iPhone set which uses a substantially different stereo baseline of $19.3$mm, the baseline model without conditioning obtains poor perceptual metrics since its predictions correspond to $\approx 63$ mm baseline and hence are spatially misaligned.
Our proposed conditioning elegantly solves this issue, allowing to predict stereoscopic content of any 3D strength (see Fig.~\ref{fig:delta-conditioning}). Consequently, our approach obtains $+3.8$ dB PSNR improvement over the non-conditioned baseline (Tab.~\ref{tab:iphone_ablation}, bottom). 
Note that our conditioning does not impact the geometric accuracy (Disp.~err), which only evaluates the relative depth ordering.

\input{src/tabs/ablation_warping}

\parsection{Warping-based versus direct} In Tab.~\ref{tab:ablation_warping}, we compare our warp-free approach to a warp-based baseline that uses DepthCrafter~\cite{depthcrafter} to estimate monocular depth. The geometric accuracy of the warping-based method directly depends on and is bounded by the accuracy of the pre-trained monocular depth method. In contrast, our approach implicitly learns a depth estimator for the direct generation task, achieving the lowest disparity errors. In Fig.~\ref{fig:qualitative_comparison_extended}, we also see that our method produces better depth ordering than warping-based alternatives. This also results in $+1.4$ dB PSNR improvement over the warp-based baseline.

\input{src/tabs/6_VAE_reconstruction_stereo4d}

\parsection{Guided Latent Decoding} We analyse the impact of our guided decoder which addresses the issue of lossy compression in VAE, as shown in Fig.~\ref{fig:vae_compression}. In Tab.~\ref{tab:ablation_study_reconstruction_stereo4d}, we evaluate our guided decoder in isolation by decoding the latents of the ground truth right image. Compared to the vanilla Stable Diffusion decoder, the guided decoder improves the PSNR by $4.1$ dB, and LPIPS score by $35\%$, by utilizing the information from the input left video. This enables better preservation of the high-frequency details, as shown in Fig.~\ref{fig:vae_compression}.

\input{src/tabs/ablation_vae}

We also evaluate the impact of the guided decoder on the end stereo conversion task in Tab.~\ref{tab:ablation_study_model_stereo4d}. When employed with our proposed warping-free approach, the guided decoder generates sharper images with high-frequency details ($16\%$ lower LPIPS, $+0.9$ dB in PSNR). It also improves stereoscopic fidelity, with an increase in $+1.3$ dB in P-PSNR and a drastic reduction of $44\%$ in Matchability error, a proxy for binocular rivalry.
Moreover, the proposed guided decoder can be used as a plug and play component, replacing the standard decoder in other frameworks. For example, replacing the decoder in M2SVid~\cite{m2svid} with ours at inference (without any specific retraining) leads to improvements in all metrics, with a particularly impressive relative improvement of $15\%$ and $34\%$ in LPIPS and Matchability error, respectively, showing the generality of the contribution. 
Finally, note that the proposed decoder does not impact the depth accuracy (Disp.~err), and leads to a small improvement in temporal stability, which can be attributed to the recovery of non-flickering high-frequency details.

\input{src/tabs/2z_2_main_SpatialVDiAVP}

\input{src/figs/2_qualitative_vertical}

\subsection{State-of-the-Art Comparison}

We benchmark our method against recent leading approaches, i.e. SVG~\cite{svg}, StereoCrafter~\cite{stereocrafter}, M2SVid~\cite{m2svid}, ReStereo\footnote{The authors of ~\cite{restereo} performed the inference for us.}~\cite{restereo}, and Eye2Eye~\cite{eye2eye}. %

\parsection{Qualitative results} As illustrated in Fig.~\ref{fig:qualitative_comparison_extended}, SVG and StereoCrafter yield blurry results. M2SVid fails to reconstruct high-frequency details (e.g., leaves and facial textures) and struggles with text (top row and Fig.~\ref{fig:teaser}). Conversely, our approach utilizes the guided-decoder to recover fine texture and texts accurately. Furthermore, Eye2Eye generates pixelated results and exhibits disparity mismatch with the ground truth due to its inherently fixed stereoscopic baseline. In contrast, our method produces aligned disparity with better depth ordering than warp-based approaches.

\parsection{Results on AVP data} We evaluate the recent methods on the AVP dataset in Tab.~\ref{tab:avp_results}. SVG, based on a pre-trained diffusion model without any task specific finetuning, obtains the worst metrics in $3/4$ categories.
The latent diffusion model based StereoCrafter struggles with detail preservation due to the lossy VAE encoding, obtaining poor PSNR score as well as Matchability error. While M2SVid partially mitigates this through image-based losses, it cannot fully address the VAE compression losses. In contrast, despite sharing the same latent diffusion backbone, our approach generates sharp images with high-frequency details, thanks to the proposed guided decoder, as evidenced by the Matchability error and P-PSNR results.
Furthermore, we observe that our direct generation approach obtains lower disparity errors compared to the warping based methods, which are inherently bounding by the limitations of the off-the-shelf monocular depth estimator.
The warp-free method Eye2Eye lacks any mechanism to control the 3D strength. Consequently, it produces spatially mis-aligned right views, leading to low overall quality metrics. We also noticed that Eye2Eye struggles to get the correct relative ordering of the objects in a scene (See Fig.~\ref{fig:qualitative_comparison_extended}), leading to higher disparity errors.
On the other hand, thanks to the proposed disparity conditioning, our approach produces pixel-aligned outputs, leading to the best overall quality metrics.  

\input{src/tabs/1c_main_stereo4d}

\parsection{Results on Stereo4D} ReStereo, which combines video generation with restoration, often hallucinates textures, resulting in poor stereoscopic fidelity on the Stereo4D dataset (Tab.~\ref{tab:stereo4d_results}). Other methods largely follow a similar trend as on the AVP set. Our method obtains the best metrics on all categories, notably achieving a $+1.5$ dB PSNR improvement over the second-best method M2SVid.

\parsection{Results on iPhone data} In Tab.~\ref{tab:iphone_results}, we report results on the iPhone dataset, which contains videos captured with substantially different 3D strength compared to our training datasets due to differences in the employed stereo cameras. Note that compared to the warp-based methods that solely perform inpainting/refinement, it is more challenging for the direct methods to synthesize right views for different stereo setups. Nevertheless, our approach obtains the best SSIM and LPIPS scores and second best PSNR score, behind only M2SVid. Conversely, Eye2Eye obtains low overall quality metrics due to lack of 3D strength control.

\input{src/tabs/2z_1_main_SpatialVDiPhone}

%% file: src/tabs/ablation_conditioning.tex
\begin{table}[b]
  \centering
  \caption{Impact of our conditioning approach (Sec.~\ref{ssec:controllable_warping_free}) on Stereo4D~\cite{stereo4d} (\textbf{top}) and iPhone Spatial Video~\cite{SpatialVD2025} (\textbf{bottom}).
  }\label{tab:iphone_ablation}
  \vspace{-3mm}
  
  \resizebox{0.48\textwidth}{!}{%
  
  \begin{tabular}{@{}l@{}c@{~~~}c@{~~~}c@{~}|@{~}c@{~~~}c@{~}|@{~}c@{~}|@{~}c@{~}}
    \toprule
    \textbf{Method}  &  \textbf{PSNR} $\uparrow$ &  \textbf{SSIM} $\uparrow$ &  \textbf{LPIPS} $\downarrow$  &  \textbf{$\mathcal{E}_\text{Match}$} $\downarrow$ &  \textbf{P-PSNR} $\uparrow$  &  \textbf{Disp.\ err} $\downarrow$ &  \textbf{Temp.\ err} $\downarrow$  \\
    \midrule

    \ours w/o Cond          &  25.1 & 0.880 & 0.192 & 28.6 & 27.2 & 1.27 & \textbf{1.28} \\
    
    \ours (ours)            &  \textbf{26.1} & \textbf{0.913} & \textbf{0.176}  & \textbf{27.8} & \textbf{27.4} & \textbf{1.24} & 1.30 \\

    \midrule

    \ours w/o Cond         & 18.7 & 0.703 & 0.289 & 30.3 & 25.2 & \textbf{0.64} & 3.23 \\
    
    \ours (ours)            & \textbf{22.5} & \textbf{0.890} & \textbf{0.193} & \textbf{26.5} & \textbf{26.2} & 0.77 & \textbf{3.10} \\
    
    \bottomrule
  \end{tabular}%
  }
  \vspace{-4mm}
\end{table}

%% file: src/tabs/ablation_warping.tex
\begin{table}[t]
  \centering
  \caption{Impact of warping free paradigm (Sec.~\ref{ssec:feed-forward}) on the Apple Vision Pro Spatial Video dataset~\cite{SpatialVD2025}. %
  }\label{tab:ablation_warping}
  \vspace{-3mm}
  
  \resizebox{0.48\textwidth}{!}{%
  
  \begin{tabular}{@{}l@{}c@{~~~}c@{~~~}c@{~}|@{~}c@{~~~}c@{~}|@{~}c@{~}|@{~}c@{~}}
    \toprule
    \textbf{Method}  &  \textbf{PSNR} $\uparrow$ &  \textbf{SSIM} $\uparrow$ &  \textbf{LPIPS} $\downarrow$  &  \textbf{$\mathcal{E}_\text{Match}$} $\downarrow$ &  \textbf{P-PSNR} $\uparrow$  &  \textbf{Disp. err} $\downarrow$ &  \textbf{Temp. err} $\downarrow$  \\
    \midrule

    Warp-based & 24.5 & 0.827 & 0.198  & \textbf{26.8} & 28.1 & 2.33 & 1.32 \\ 
    
    \ours (ours)     & \textbf{25.9} & \textbf{0.894} & \textbf{0.196} & 30.9 & \textbf{28.4} & \textbf{1.74} & \textbf{1.31} \\
    
    \bottomrule
  \end{tabular}%
  }
\end{table}

%% file: src/tabs/6_VAE_reconstruction_stereo4d.tex
\begin{table}[t]
  \centering
  \caption{Reconstruction results on Stereo4D~\cite{stereo4d}, where the VAE is applied on the ground truth right views.}
  \vspace{-3mm}
  \label{tab:ablation_study_reconstruction_stereo4d}
  
  \resizebox{0.38\textwidth}{!}{%
  \begin{tabular}{l ccc}
    \toprule
    \textbf{Model} & \textbf{PSNR} $\uparrow$ & \textbf{SSIM} $\uparrow$ & \textbf{LPIPS} $\downarrow$ \\
    \midrule

    Stable Diffusion  & 30.2 & 0.963 & 0.106  \\
    Guided Decoder (Ours)        & \textbf{34.3} & \textbf{0.981} & \textbf{0.068} \\

    \bottomrule
  \end{tabular}%
  }
\end{table}

%% file: src/tabs/ablation_vae.tex
\begin{table}[t]
  \centering
  \caption{Impact of our guided-VAE decoder $\mathcal{D'}$ (Sec.~\ref{ssec:guided_decoding}) for mono-to-stereo conversion on the Stereo4d dataset~\cite{stereo4d}.}
  \label{tab:ablation_study_model_stereo4d}
  \vspace{-3mm}
  \resizebox{0.48\textwidth}{!}{%
  \begin{tabular}{@{}l@{}c@{~~~}c@{~~~}c@{~}|@{~}c@{~~~}c@{~}|@{~}c@{~}|@{~}c@{~}}
    \toprule
    \textbf{Model} & \textbf{PSNR} $\uparrow$ & \textbf{SSIM} $\uparrow$ & \textbf{LPIPS} $\downarrow$  &\textbf{$\mathcal{E}_\text{Match}$}  $\downarrow$& \textbf{P-PSNR} $\uparrow$ & \textbf{Disp. err} $\downarrow$ & \textbf{Temp. err} $\downarrow$\\
    \midrule

    \ours w/o $\mathcal{D'}$  & 25.2 & 0.895 & 0.212  & 41.9 & 26.1 & \textbf{1.23} & 1.37 \\

    \ours (ours) &  \textbf{26.1} & \textbf{0.913} & \textbf{0.176} & \textbf{27.8} & \textbf{27.4} & 1.24 & \textbf{1.30}
    \\ \midrule
    
    M2SVid &   24.6  &  0.819  &  0.206  &  39.6  &  26.3   &  \textbf{1.56}  &  1.35  \\
    M2SVid + $\mathcal{D'}$ & \textbf{25.2} & \textbf{0.832} & \textbf{0.175}  &  \textbf{24.8} & \textbf{27.5} & 1.61 & \textbf{1.27} \\
    \bottomrule
  \end{tabular}%
  }
  \vspace{-4mm}
\end{table}

%% file: src/tabs/2z_2_main_SpatialVDiAVP.tex
\begin{table}[t]
  \centering
  
  \caption{State-of-the-art comparison on the Apple Vision Pro Spatial Video dataset~\cite{SpatialVD2025}. The baseline of the dataset is similar to during training while the content is out-of-distribution.
  }\label{tab:avp_results}
  \vspace{-3mm}
  
  \resizebox{0.48\textwidth}{!}{%
  
  \begin{tabular}{@{}l@{~~}c@{~~~}c@{~~~}c|c@{~~~}c|c|c@{~}}
    \toprule
    \textbf{Method}  &  \textbf{PSNR} $\uparrow$ &  \textbf{SSIM} $\uparrow$ &  \textbf{LPIPS} $\downarrow$  &  \textbf{$\mathcal{E}_\text{Match}$} $\downarrow$ &  \textbf{P-PSNR} $\uparrow$  &  \textbf{Disp. err} $\downarrow$ &  \textbf{Temp. err} $\downarrow$  \\
    \midrule
    
    SVG                    & 19.3 & 0.690 & 0.410 & 56.3 & 20.2 & 3.71 & 8.49 \\
    StereoCrafter & 22.5 & 0.826 & 0.323 & 51.8 & 22.6 & 2.30 & 1.71 \\

    M2SVid                & 24.4 & 0.821 & 0.221 & 41.5 & 27.3 & 2.30 & 1.35 \\
    
    Eye2Eye            & 20.6 & 0.733 & 0.392 & 39.2 & 23.9 & 3.82 & 2.18 \\
    \rowcolor{almond}  \ours     & \textbf{25.9} & \textbf{0.894} & \textbf{0.196} & \textbf{30.9} & \textbf{28.4} & \textbf{1.74} & \textbf{1.31} \\

    \bottomrule
  \end{tabular}%
  }
  \vspace{-4mm}
\end{table}

%% file: src/figs/2_qualitative_vertical.tex
\begin{figure*}[t!]
  \centering
    \includegraphics[width=\textwidth]{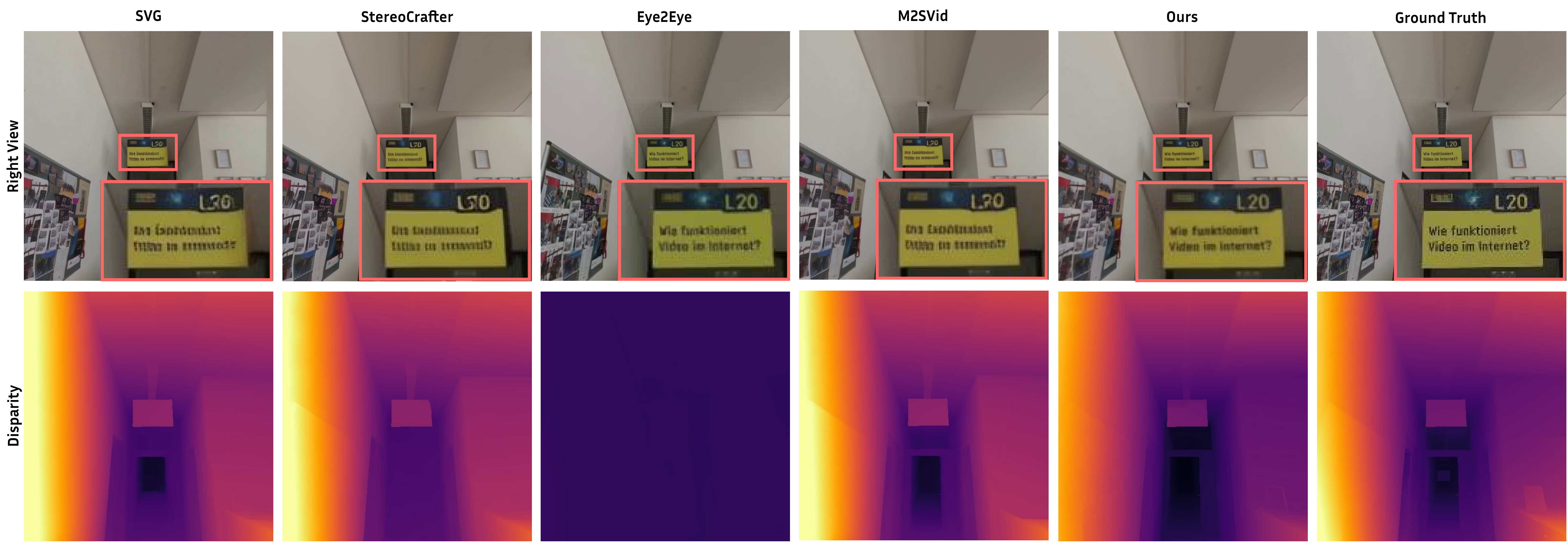}
    \includegraphics[width=\textwidth]{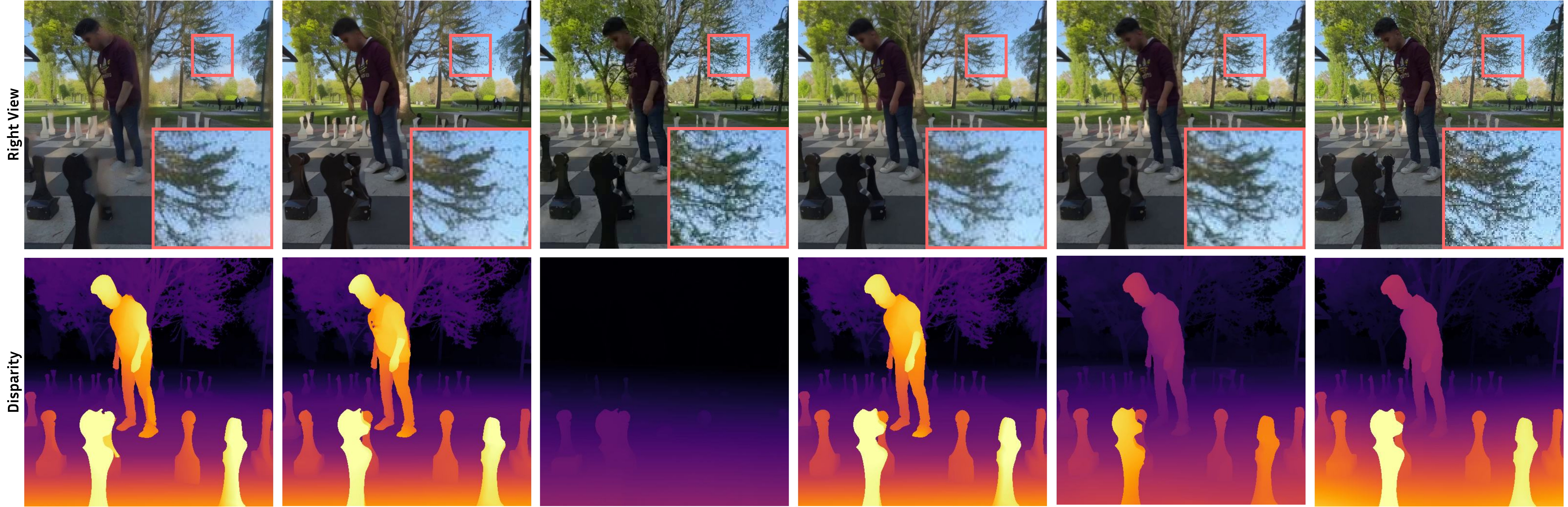}
    \vspace{-7mm}
      \caption{
    \textbf{Qualitative comparison.}
    Our method generates better textures with high-frequency details. The geometry of the generated stereo sample is also more correct -- in line with the ground truth disparity.
    \textit{Note: This figure has been compressed for arXiv submission. Please refer to the project page for the uncompressed version.}
  }\label{fig:qualitative_comparison_extended}
    \vspace{-4mm}
  
\end{figure*}

%% file: src/tabs/1c_main_stereo4d.tex
\begin{table}[b]
    \centering
    \vspace{-4mm}
    \caption{State-of-the-art comparison on  the  Stereo4D~\cite{stereo4d}  test set. 
    }
    \vspace{-3mm}
    \label{tab:stereo4d_results}
    
    \resizebox{0.48\textwidth}{!}{%
    \begin{tabular}{@{}l@{~~}c@{~~~}c@{~~~}c|c@{~~~}c|c|c@{~}}
        \toprule
        \textbf{Method}  &  \textbf{PSNR} $\uparrow$ &  \textbf{SSIM} $\uparrow$ &  \textbf{LPIPS} $\downarrow$  &  \textbf{$\mathcal{E}_\text{Match}$} $\downarrow$ &  \textbf{P-PSNR} $\uparrow$  &  \textbf{Disp. err} $\downarrow$ &  \textbf{Temp. err} $\downarrow$  \\
        \midrule
        
        SVG                     &  22.9  &  0.874  &  0.258  &  40.0  &  23.2  &  1.69  &  1.45  \\
        StereoCrafter                &  23.6  &  0.874  &  0.258  &  43.9  &  22.8  &  1.75  &  1.42  \\
        M2SVid                &  24.6  &  0.819  &  0.206  &  39.6  &  26.3  &  1.56  &  1.35  \\
        Eye2Eye              &  21.1  &  0.780  &  0.313  &  35.8  &  23.0  &  2.45  &  2.11   \\
        ReStereo            &  21.2  &  0.788  &  0.307  &  41.3  &  23.0  &  1.95  &  1.52  \\

    	 \rowcolor{almond}  \ours  &  \textbf{26.1} &  \textbf{0.913}  &\textbf{0.176}  &  \textbf{27.8}  &  \textbf{27.4}   &  \textbf{1.24}  &  \textbf{1.30}  \\

        \bottomrule
    \end{tabular}%
    }
    
\end{table}

%% file: src/tabs/2z_1_main_SpatialVDiPhone.tex
\begin{table}[b]
  \centering
  \vspace{-4mm}
  \caption{State-of-the-art comparison on the iPhone portion of Spatial Video dataset~\cite{SpatialVD2025}. Both calibration and video content are out-of-distribution.
  }\label{tab:iphone_results}
  \vspace{-3mm}
  
  \resizebox{0.48\textwidth}{!}{%
    \setlength{\aboverulesep}{0pt}
    \setlength{\belowrulesep}{0pt}
    \renewcommand{\arraystretch}{1.15}
  \begin{tabular}{@{}l@{~~}c@{~~~}c@{~~~}c|c@{~~~}c|c|c@{~}}
    \toprule
    \textbf{Method}  &  \textbf{PSNR} $\uparrow$ &  \textbf{SSIM} $\uparrow$ &  \textbf{LPIPS} $\downarrow$  &  \textbf{$\mathcal{E}_\text{Match}$} $\downarrow$ &  \textbf{P-PSNR} $\uparrow$  &  \textbf{Disp. err} $\downarrow$ &  \textbf{Temp.  err} $\downarrow$  \\
    \midrule
    
    SVG                     & 16.3 & 0.600 & 0.387 & 49.0 & 17.8  & 1.45 & 13.2 \\
    StereoCrafter           & 21.9 & 0.877 & 0.257 & 45.1 & 21.4  & 0.63 & 3.43 \\
    M2SVid                  & \textbf{22.9} & 0.865 & 0.205 & 38.4 & 25.1 & \textbf{0.60} & 3.06 \\
    Eye2Eye                & 20.2 & 0.818 & 0.281 & 32.0 & 22.8 & 1.11 & \textbf{3.02} \\
    \rowcolor{almond}  \ours      & 22.5 & \textbf{0.890} & \textbf{0.193} & \textbf{26.5} & \textbf{26.2} & 0.77 & 3.10 \\

    \bottomrule
  \end{tabular}%
  }
\end{table}

%% file: sec/7_conclusion.tex
\section{Conclusion}
\label{sec:conclusion}

We present a 3D-controllable, warping-free framework for converting monocular videos into high-fidelity binocular stereo videos. By integrating a novel guided VAE decoder into a disparity-conditioned latent diffusion architecture, our method ensures sharp, epipolar-consistent output while enabling intuitive scalar control over disparity, setting a new state-of-the-art across diverse datasets.

%% file: sec/X_suppl.tex
\clearpage
\setcounter{page}{1}
\maketitlesupplementary

\appendix

In Sec.~\ref{sec:supp_implementation}, we provide implementation details for our approach. Then, in Sec~\ref{sec:supp_attention}, we explain in detail our epipolar-guided decoder cross-attention (see Sec.~4.3 of the main paper). We follow with a short background on diffusion models and a derivation of our 1-step feed-forward model in Sec.~\ref{sec:supp_diffusion}. Next, we provide more details on the evaluation protocol in Sec.~\ref{sec:eval-protocol-details}. In particular, we give intuitive explanations and insights on the proposed metrics (see Sec.~5 of the main paper).
In Sec.~\ref{sec:results-sup}, we provide additional results for our method. In particular, we conduct a user study on headset to validate that our superiority according to metrics translates to a better user experience on device. Notably, we also provide latency comparisons between our method and state-of-the-art methods.
In Sec.~\ref{sec:results-sup}, we provide additional results for our method. In particular, we conduct a user study on a headset to validate that our superiority according to metrics translates to a better user on-device experience. Notably, we also provide latency comparisons between our method and state-of-the-art methods.

\vspace{0.5em}
\noindent \textbf{Video Results.} We strongly encourage the reader to also view the accompanying \url{https://elastic3d.github.io/}. This interactive viewer contains more extensive video comparisons.

\section{Implementation Details}
\label{sec:supp_implementation}

Our training pipeline consists of two distinct stages: first, the training of the warping-free synthesis core (U-Net), and second, the training of the Guided Decoder. For both stages, we utilize the Stereo4D and Ego4D datasets.

\subsection{Training Data Handling}
Following~\cite{m2svid}, we use a batch size of 1 and apply random temporal sub-sampling to each sample to achieve a target frame rate between 5 and 30 FPS. Additionally, we apply random spatial scaling with factors in $[0.3, 1.0]$ and employ a resolution and frame length bucketing strategy for data augmentation, as detailed in Table~\ref{tab:bucketing}.

\paragraph{Rectification quality of training data.}
Following~\cite{m2svid}, we estimate rectifying homographies using a standard SfM-style robust pipeline based on LoFTR matching followed by RANSAC. After rectification, the residual vertical parallax in the training pairs is below 2 pixels. This ensures that the stereo supervision is sufficiently well aligned for learning both disparity control and the guided decoding mechanism.

\paragraph{Synthetic Baseline Augmentation.} To improve the model's robustness to varying disparity ranges, we generate a synthetic subset of data. We pre-compute 2,500 random samples from Stereo4D and create synthetic stereo pairs by forward-warping the left view pixels using the ground truth depth, scaled by a random factor $s$. The corresponding conditioning scalar is adjusted to $s \cdot \delta$. We employ the following discrete set of scaling factors: $s \in \{0.05, 0.1, 0.2, 0.4, 0.6, 0.8, 1.25, 1.5, 2.0, 3.0\}$.

\paragraph{Zero-Disparity Augmentation.} For $1\%$ of the training batches, we employ a zero-disparity augmentation where the target frame is identical to the input frame, and the disparity conditioning is explicitly set to $\delta=0$.

\begin{table}[tb]
    \centering
    \small
    \caption{\textbf{Training Bucketing Strategy.} Combinations of resolution and frame counts used for augmentation.}
    \label{tab:bucketing}
    \resizebox{0.41\textwidth}{!}{%
    \begin{tabular}{lcc}
        \toprule
        \textbf{Resolution} $[H, W]$ & \textbf{Frames} $N$ & \textbf{Aspect Ratio} \\
        \midrule
        $[512, 512]$ & 4 & $1:1$ \\
        $[768, 320]$, $[320, 768]$ & 4 & $2.4:1$ \\
        $[1024, 256]$, $[256, 1024]$ & 4 & $4:1$ \\
        $[1280, 256]$, $[256, 1280]$ & 3 & $5:1$ \\
        $[256, 256]$ & 16 & $1:1$ \\
        $[192, 192]$ & 25 & $1:1$ \\
        $[128, 256]$, $[256, 128]$ & 25 & $2:1$ \\
        \bottomrule
    \end{tabular}
    }
\end{table}

\subsection{U-Net Training}

We utilize Stable Video Diffusion (SVD) as our backbone, freezing the VAE and training the U-Net for the 1-step latent synthesis task.

\paragraph{Optimization Framework.} To manage the memory footprint during training, we utilize DeepSpeed Stage 2~\cite{rasley2020deepspeed} with FP16 mixed precision. Furthermore, we optimize memory usage by offloading optimizer states to the CPU.

\paragraph{Loss and Hyperparameters.} We employ a composite loss function consisting of an $L_2$ loss in the latent space, combined with pixel-space losses computed via the frozen Stable Diffusion VAE Decoder. The pixel-space objectives include $L_1$, SSIM, and LPIPS. We train with a learning rate of $2 \times 10^{-6}$.

\paragraph{Training Phases and Data Sampling.} The training is performed in two distinct phases. First, we train a base unconditional model for $200\text{k}$ iterations. Subsequently, we fine-tune the model for an additional $170\text{k}$ iterations with the disparity conditioning mechanism enabled.
During this second phase each batch is sampled with equal probability ($1/3$) from three sources: the standard Stereo4D dataset, the Ego4D dataset, and the Synthetic Stereo4D subset described above.

\paragraph{Conditioning Mechanism.} The scalar median disparity $\delta$ is projected into a high-dimensional token embedding $\tau(\delta)$. This projection is achieved by broadcasting the scalar value across the vector until it matches the required token dimensionality. This token is then concatenated with the standard CLIP embeddings of the input frame and injected into the U-Net. This strategy requires no additional learnable parameters. We found that scaling the raw disparities by a factor of $\approx 10^{-2}$ is crucial for numerical stability.

\subsection{Guided Decoder Training}

The Guided Decoder is trained separately to reconstruct the ground-truth right view $V_R$ from its latent $z_R$, using the left video $V_L$ as guidance.

\paragraph{Precision and Optimization.} To avoid overflow in the introduced attention, we found that training in \texttt{bfloat16} precision is crucial. Following~\cite{SplatDiff}, we optimize the decoder using an equally weighted reconstruction loss combining $L_1$ and LPIPS objectives, with a learning rate of $1 \times 10^{-4}$. We train for 50k iterations.

\section{Epipolar Guided Decoder Cross-Attention}
\label{sec:supp_attention}

In this section, we provide the formal definition of the epipolar cross-attention mechanism~\cite{epipolartransformer} within the guided VAE decoder. In short, the employed attention mechanism is a standard cross attention implementation~\cite{vaswani2017attention} that chooses the query, key, and value locations according to epipolar geometry.

Let $h_i \in \mathbb{R}^{H_i \times W_i \times C}$ be the intermediate feature map in the decoder at layer $i$, and $g_i \in \mathbb{R}^{H_i \times W_i \times C}$ be the corresponding guidance feature map extracted from the input left view $V_L$.

For a specific pixel location $p=(u, v)$, we first compute the query vector $Q_p \in \mathbb{R}^d$ by projecting the decoder feature $h_i(p)$ using a learnable weight matrix $\mathcal{W}_Q \in \mathbb{R}^{C \times d}$:

\begin{equation}
    Q_p = h_i(p) \mathcal{W}_Q
\end{equation}

Similarly, for the guidance features, we restrict our scope to the corresponding horizontal scanline (epipolar line) $v$. We compute the keys $K_{row} \in \mathbb{R}^{W_i \times d}$ and values $V_{row} \in \mathbb{R}^{W_i \times d}$ by projecting the entire row of guidance features $g_i(v) \in \mathbb{R}^{W_i \times C}$ via learnable matrices $\mathcal{W}_K, \mathcal{W}_V \in \mathbb{R}^{C \times d}$:
\begin{equation}
    K_{row} = g_i(v) \mathcal{W}_K, \quad V_{row} = g_i(v) \mathcal{W}_V
\end{equation}

We then calculate the intermediate attention result $\alpha_p \in \mathbb{R}^d$ via scaled dot-product attention over the row:
\begin{equation}
    \alpha_p = \text{Softmax}\left( \frac{Q_p K_{row}^\top}{\sqrt{d}} \right) V_{row}
\end{equation}

The full epipolar attention module $\mathcal{A}_{\text{epipolar}}$ is obtained by projecting this result back to the original channel dimension $C$ using a linear output projection $\mathcal{W}_{\text{out}} \in \mathbb{R}^{d \times C}$:
\begin{equation}
    \mathcal{A}_{\text{epipolar}}(h_i(p), g_i) = \alpha_p \mathcal{W}_{\text{out}}
\end{equation}

Finally, the decoder features are updated via a residual connection:
\begin{equation}
    h'_i(p) = h_i(p) + \mathcal{A}_{\text{epipolar}}(h_i(p), g_i)
\end{equation}

\paragraph{Zero-Initialization.} To ensure the Guided Decoder preserves the identity mapping of the pre-trained decoder at the beginning of training, we initialize the weights of the linear output projection layer $\mathcal{W}_{\text{out}}$ (inside $\mathcal{A}_{\text{epipolar}}$) to zero.

\section{Background on 1-Step Diffusion Formulation}
\label{sec:supp_diffusion}

Our synthesis core utilizes the diffusion backbone $f_\theta$ for 1-step inference, transforming it into an efficient feed-forward generator. We here provide background on diffusion and derive the formulation of our 1-step inference approach.

\parsection{DDPMs}
The core mechanism of DDPMs~\cite{ho2020denoising} involves learning to reverse a stochastic forward chain that gradually destroys data structure. This forward process maps a data distribution $p_0$ to a Gaussian noise distribution $p_T$ across discrete timesteps $t = 1, \dots, T$. By introducing Gaussian noise with variance $\beta_t$ at each step, the transition can be expressed in closed form as $\mathbf{x}_t = \sqrt{\bar\alpha_t}\mathbf{x}_0 + \sqrt{1 - \bar\alpha_t} \boldsymbol{\epsilon}$. Here, $\mathbf{x}_0$ represents the clean data sample, $\boldsymbol{\epsilon} \sim \mathcal{N}(\mathbf{0}, \mathbf{I})$ denotes the noise term, $\alpha_t = 1 - \beta_t$, and $\bar\alpha_t = \prod_{\tau=1}^t \alpha_\tau$. A parameterized denoising network $\hat{\mathbf{v}}_{\theta}(\mathbf{x}_t, t)$ is trained to approximate the reverse transition, effectively predicting $\mathbf{x}_{t-1}$ given the noisy state $\mathbf{x}_t$ to iteratively recover the original signal.

\parsection{DDIMs}
To mitigate the computational cost associated with the Markovian sampling of DDPMs, Song~\textit{et al.}~\cite{song2021denoising} proposed Denoising Diffusion Implicit Models (DDIMs). By reformulating the diffusion as a non-Markovian process, DDIMs maintain the original training objective but facilitate deterministic sampling. This permits the generation of high-fidelity samples in significantly fewer steps (typically 25 to 50) compared to the full trajectory required by DDPMs.

\parsection{LDMs}
Directly applying diffusion models to high-resolution pixel space is computationally expensive. Latent Diffusion Models (LDMs)~\cite{rombach2022high} address this by shifting the generative process into the compressed latent space of a Variational Autoencoder (VAE)~\cite{kingma2013auto}. The VAE comprises an encoder $E$ and a decoder $D$ such that $D(E(\mathbf{x})) \approx \mathbf{x}$. During the training of the diffusion backbone $\hat{\mathbf{v}}_{\theta}$, the VAE weights are frozen, allowing the model to learn the data distribution within a lower-dimensional space. This approach preserves the semantic structure of the data while substantially reducing computational complexity.

\parsection{Conditional diffusion models}
Recent advancements~\cite{saharia2022photorealistic, zhang2023adding} have extended the capabilities of diffusion models by incorporating conditional inputs. The denoising network is modified to the form $\hat{\mathbf{v}}_{\theta}(\mathbf{x}_t, t, c)$, where $c$ represents auxiliary information. This conditioning signal enables control over the generation process using diverse modalities, including text prompts~\cite{saharia2022photorealistic}, reference images, or spatial guidance such as depth maps and pose estimations~\cite{zhang2023adding}.

\parsection{Training and inference with conditional LDMs}
In the training phase, a data sample $\mathbf{x}$ (e.g., an image) is processed by encoding $\mathbf{x}$ into a latent representation $\mathbf{z} = E(\mathbf{x})$. This latent vector is then perturbed via the forward diffusion process:
\[
\mathbf{z}_t = \sqrt{\bar\alpha_t} \mathbf{z} + \sqrt{1 - \bar\alpha_t} \boldsymbol{\epsilon},
\]
where $\boldsymbol{\epsilon} \sim \mathcal{N}(\mathbf{0}, \mathbf{I})$ and $\bar\alpha_t$ is determined by the noise schedule. Modern implementations often utilize \textit{v-parameterization}~\cite{salimans2022progressive} for improved stability. Instead of directly predicting the noise, the model $\hat{\mathbf{v}}_{\theta}(\mathbf{z}_t, t, c)$ learns to regress a target velocity $\mathbf{v}$, defined as:
\[
\mathbf{v} = \alpha_t \boldsymbol{\epsilon} - \sqrt{1 - \bar\alpha_t} \mathbf{z}.
\]
Consequently, the network—typically a U-Net~\cite{ronneberger2015u}—is optimized by minimizing the following reconstruction objective:
\begin{equation}
\label{eq:diffusion_loss}
    \mathcal{L} = \mathbb{E}_{\mathbf{z}, c, \boldsymbol{\epsilon}, t} \left[ \| \mathbf{v} - \hat{\mathbf{v}}_{\theta}(\mathbf{z}_t, t, c) \|^2 \right].
\end{equation}
During inference, the system reconstructs the sample by iteratively denoising a pure noise tensor $\mathbf{z}_T$ over $T$ steps, guided by the learned denoiser $\hat{\mathbf{v}}_{\theta}$ and the condition $c$.

\parsection{1-step diffusion as feed-forward models}
Garcia~\textit{et al.}~\cite{marigolde2e} propose a method to adapt pre-trained diffusion U-Nets as deterministic feed-forward models for pixel-to-pixel tasks. This is achieved by fixing the sampling timestep to $t=T$ and replacing the stochastic noise component $\boldsymbol{\epsilon}$ with its expectation (zero). Under the assumption that $t=T$, we have $\bar\alpha_T \approx 0$, which simplifies the forward process terms to:
\[
\mathbf{z}_T = \sqrt{\bar\alpha_T} \mathbf{z} + \sqrt{1 - \bar\alpha_T} \boldsymbol{\epsilon} \approx \mathbf{0}
\]
and
\[
\mathbf{v} = \alpha_t \boldsymbol{\epsilon} - \sqrt{1 - \bar\alpha_t}  \mathbf{z} \approx - \mathbf{z}.
\]
In this regime, the model learns to predict the clean latent structure directly from a zero-initialized input $\mathbf{z}_T = \mathbf{0}$. The standard diffusion loss in Equation~\ref{eq:diffusion_loss} effectively converges to an $L_2$ regression between the ground truth $-\mathbf{z}$ and the prediction $\hat{\mathbf{v}}_{\theta}(\mathbf{0}, T, c)$:
\begin{equation}\label{eq:latent}
    \mathcal{L}_{latent} = \mathbb{E}_{\mathbf{z}, c} \left[ \| (-\mathbf{z}) - \hat{\mathbf{v}}_{\theta}(\mathbf{0}, T, c) \|^2 \right].
\end{equation}
Rather than optimizing in latent space, Garcia~\textit{et al.} decode the predicted latents $\mathbf{\hat{z}} = -\hat{\mathbf{v}}_{\theta}(\mathbf{0}, T, c)$ via the decoder $D$ to obtain a reconstruction $\mathbf{\hat{x}} = D(\mathbf{\hat{z}})$. This enables the optimization of a task-specific loss $\mathcal{L} = L_{task}(\mathbf{x}, \mathbf{\hat{x}})$ directly in the image domain.

\parsection{Our own formulation} Our synthesis core utilizes the diffusion backbone $f_\theta$ for \textbf{1-step inference}, transforming it into an efficient feed-forward generator. While the original diffusion model predicts the noise $\epsilon$ iteratively, we train $f_\theta$ to directly predict the clean latent data $\hat{z}_R$ from a zero-noise input $\mathbf{0}$ and a fixed maximum timestep (or pseudo-timestep $T$), conditioned on the input $z_L$ and the disparity token $\tau(\delta)$:
\begin{equation}
   \hat{z}_R = f_\theta(\mathbf{0}, T, z_L, \tau(\delta)) = - \hat{\mathbf{v}}_\theta(\mathbf{0}, T, c)
\end{equation}
where we have simplified $f_\theta = - \hat{\mathbf{v}}_\theta$.

This approach utilizes the rich video and image priors learned during SVD pre-training but bypasses the computationally expensive iterative scheduler. By employing a composite loss (L2 latent + pixel-space L1/SSIM/LPIPS) to minimize the distance between $\hat{z}_R$ and the ground truth $z_R$, we effectively train the U-Net as a powerful image-to-image translation network for direct stereo synthesis.

\section{More Details on Evaluation Protocol}
\label{sec:eval-protocol-details}

First, in Sec.~\ref{supp:metric_sensitivity}, we analyze the robustness of our proposed metrics. Next, we provide an intuitive explanation of the Matchability Error and what it measures in Sec.~\ref{supp:matchability}.

\subsection{Standardization of Warping-Based Baselines}
\label{sec:baseline_standardization}

Warping-based methods depend on monocular depth predictions whose scale is inherently ambiguous. To ensure fair comparison, we align the predicted disparity of each baseline to the ground-truth disparity of each target video using an L1 scale-and-shift fit. This standardizes the global stereo strength across methods. We do not clip the maximum disparity; however, we enforce positive disparities to maintain valid rectified geometry.

\subsection{Correlation with Human Preference}
\label{sec:sps_correlation}

Beyond sensitivity to controlled degradations, we also verify whether the proposed metrics align with human judgments. To this end, we correlate the objective scores with the subjective preference score collected in our pairwise user study:
\begin{equation}
\mathrm{SPS} = \frac{N_{\text{Ours}} + 0.5 \times N_{\text{Tie}}}{N_{\text{Total}}}.
\end{equation}

\begin{figure*}[t]
  \centering
  \includegraphics[width=0.75\linewidth]{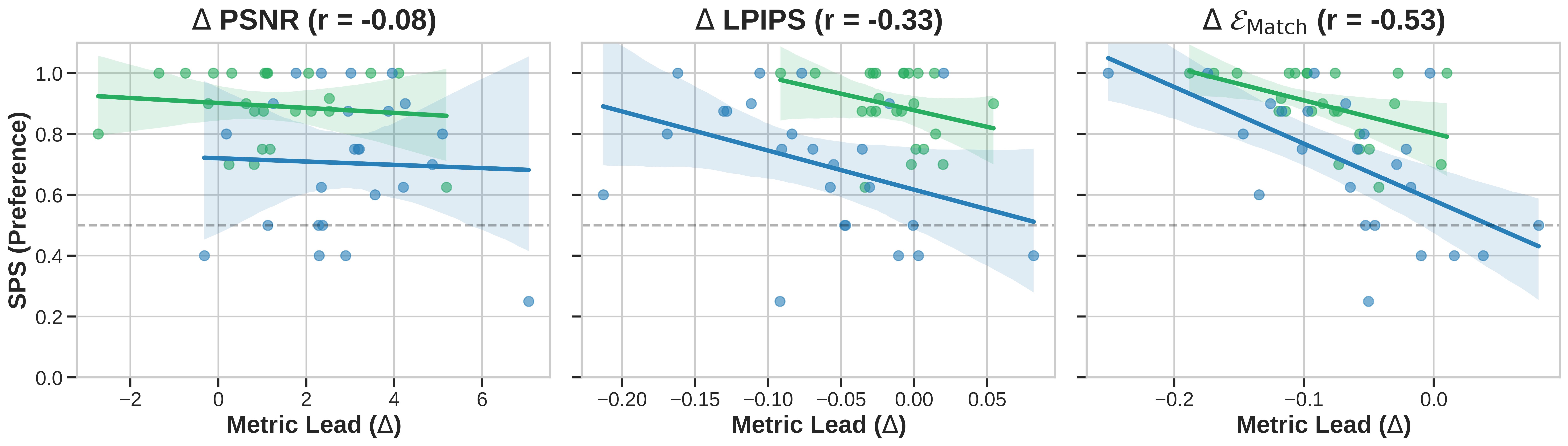}
  \caption{\textbf{Correlation with human preference.} 
  Our proposed Matchability Error aligns better with subjective preference scores than conventional metrics. \textcolor{green}{Green}: Ours vs.\ M2SVid. \textcolor{blue}{Blue}: Ours vs.\ Eye2Eye. Lower Matchability Error corresponds to higher user preference.}
  \label{fig:sps_corr}
\end{figure*}

Figure~\ref{fig:sps_corr} shows that our proposed Matchability Error $\mathcal{E}_{\text{Match}}$ correlates substantially better with human preference than conventional full-reference metrics. In particular, we observe a negative correlation of $r=-0.53$, indicating that users consistently prefer results with lower Matchability Error. In contrast, $\Delta$LPIPS exhibits a weaker correlation ($r=-0.33$), while $\Delta$PSNR shows almost no correlation with human preference ($r=-0.08$). We hypothesize that this is because PSNR can favor overly smooth or blurry results, which may improve pixel-wise agreement while degrading perceived stereoscopic fidelity.

This supports our claim that Matchability Error better captures the aspects of stereoscopic fidelity that matter perceptually in our setting, especially the preservation of sharp, matchable, and geometrically plausible details. In contrast, conventional metrics are less informative for comparing modern stereo video synthesis methods, where slight spatial deviations and alternative plausible textures can strongly affect pixel-wise scores without necessarily degrading the perceived 3D viewing experience.

\subsection{Metric Sensitivity Analysis}
\label{supp:metric_sensitivity}

In the main paper, we introduced two component-wise metrics: Matchability Error ($\mathcal{E}_{\text{Match}}$) and Patch-wise PSNR (P-PSNR). These were designed to complement standard full-reference metrics (SSIM, PSNR) by decoupling the evaluation of \textit{stereoscopic fidelity} (texture preservation) from \textit{geometric correctness} and \textit{3D-strength controllability} (3D alignment).

To validate these design choices and demonstrate the robustness of our proposed metrics, we conducted a sensitivity analysis using the Stereo4D dataset. We took the predictions generated by our model and applied two distinct post-processing degradations before computing the metrics:
\begin{enumerate}
    \item \textbf{Horizontal Shifting:} We artificially shift the right view horizontally by $N$ pixels. This simulates geometric rectification errors or incorrect 3D strength (disparity) predictions.
    \item \textbf{Gaussian Blur:} We apply a Gaussian blur with varying $\sigma$. This simulates texture loss and over-smoothing, which are common artifacts in latent diffusion models.
\end{enumerate}

\subsubsection{Sensitivity to Geometric Misalignment}

\begin{figure}[t]
    \centering
    \includegraphics[width=1.0\linewidth]{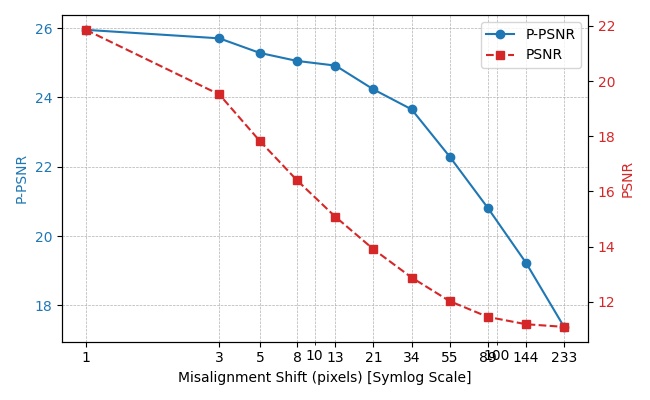}
    \caption{\textbf{Robustness to Misalignment.} Sensitivity of Standard PSNR vs. our PatchPSNR (P-PSNR) to horizontal pixel shifts. P-PSNR remains robust to small geometric errors, focusing on texture quality.}
    \label{fig:shift_psnr_vs_patchpsnr}
\end{figure}

Standard pixel-wise metrics are heavily penalized by slight spatial misalignments, even if the textural content is perfectly preserved.
In Fig.~\ref{fig:shift_psnr_vs_patchpsnr}, we observe that standard PSNR drops precipitously with even minor pixel shifts. In contrast, our P-PSNR exhibits high robustness, maintaining a relatively good score for shifts up to $\sim13$ pixels. This shows that P-PSNR disentangles photometric consistency from geometric alignment, effectively finding the best matching patches along the epipolar line.

\begin{figure}[t]
    \centering
    \includegraphics[width=1.0\linewidth]{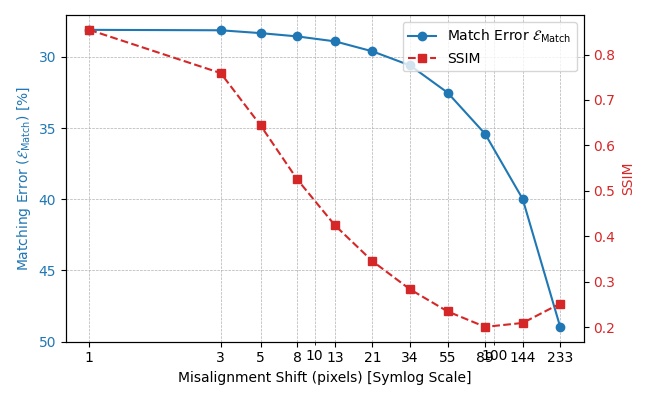}
    \caption{\textbf{Robustness to Misalignment.} Sensitivity of SSIM vs. our Matchability Error to horizontal pixel shifts. Matchability remains stable, effectively decoupling texture evaluation from geometry.}
    \label{fig:shift_ssim_vs_match_error}
\end{figure}

Similarly, in Fig.~\ref{fig:shift_ssim_vs_match_error}, SSIM degrades rapidly as the shift increases. Conversely, the Matchability Error remains largely stable regardless of the horizontal shift. This is expected, as the feature matcher (DeDoDe~\cite{dedodev2}) is designed to be invariant to translation.
This stability ensures that we are measuring the \textit{presence} of matchable texture details, rather than their precise pixel location, which is evaluated separately by our geometric correctness metric (Disp.~err).

\subsubsection{Sensitivity to Texture Degradation}

\begin{figure}[t]
    \centering
    \includegraphics[width=1.0\linewidth]{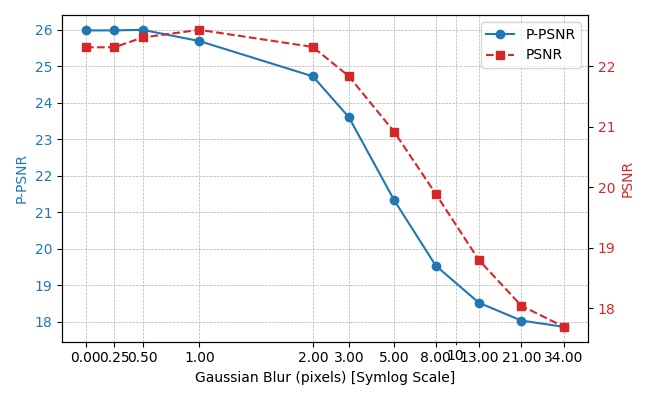}
    \caption{\textbf{Sensitivity to Blur.} Sensitivity of Standard PSNR vs. our PatchPSNR (P-PSNR) to Gaussian blur. Both metrics follow a similar trend, validating P-PSNR as a reliable quality estimator.}
    \label{fig:blur_psnr_vs_patchpsnr}
\end{figure}

While robustness to shift is desirable, the metrics must still remain sensitive to image degradation (blur).
In Fig.~\ref{fig:blur_psnr_vs_patchpsnr}, we compare the sensitivity to Gaussian blur.
We observe that P-PSNR follows the same trend as standard PSNR, confirming that it remains a valid proxy for image quality despite its spatial relaxation.
Interestingly, we observe the common phenomenon where a slight blur ($\sigma=1.0$) results in a marginal increase in PSNR ($+0.4$ dB).

\begin{figure}[t]
    \centering
    \includegraphics[width=1.0\linewidth]{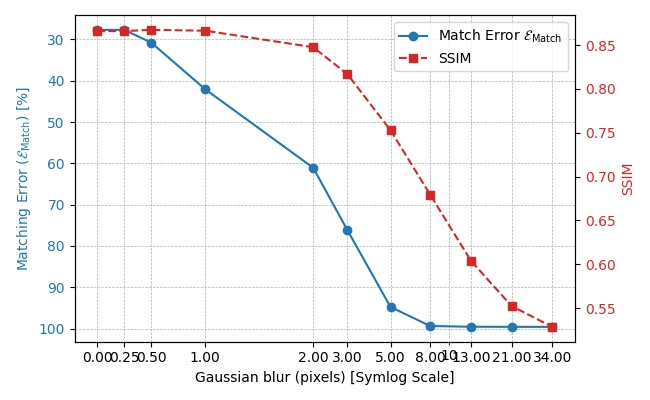}
    \caption{\textbf{Sensitivity to Blur.} Sensitivity of SSIM vs. our Matchability Error to Gaussian blur. Matchability is more sensitive to blur than SSIM, effectively penalizing over-smoothed generations.}
    \label{fig:blur_ssim_vs_match_error}
\end{figure}

Crucially, Fig.~\ref{fig:blur_ssim_vs_match_error} demonstrates the superiority of Matchability Error in detecting over-smoothing. While SSIM decays relatively slowly with increased blur, the Matchability Error rises sharply. This indicates that our metric is significantly more sensitive to the loss of high-frequency details than SSIM.
A blurry image might still look ``structurally" similar (high SSIM), but it will fail to produce reliable keypoints, leading to a high Matchability Error.
This makes it an effective metric for detecting the "waxy" or "washed-out" artifacts often produced by video diffusion models.

\begin{figure*}[t]
    \centering
    \includegraphics[width=\linewidth]{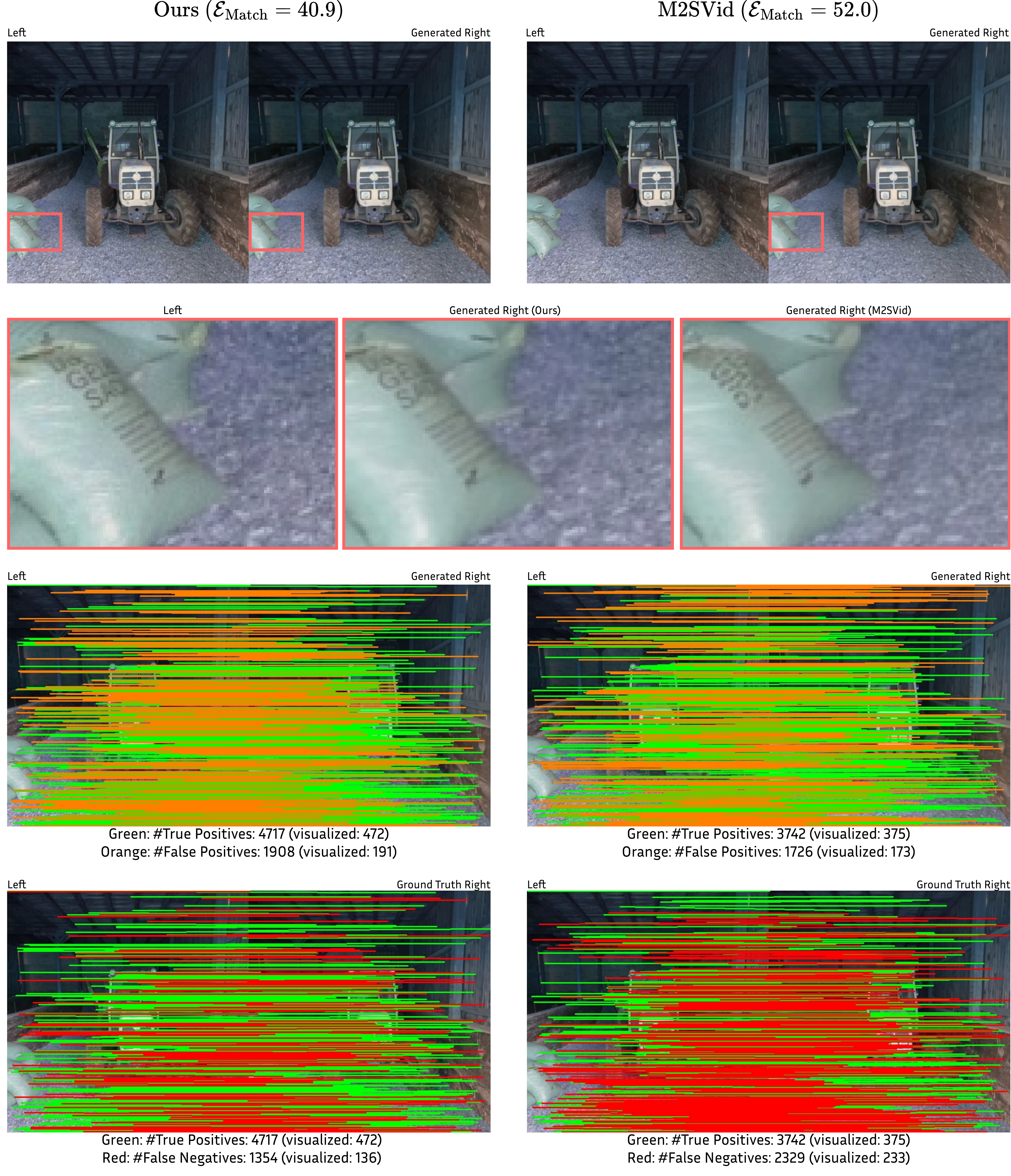}
    \caption{\textbf{Visualization of Matchability Error Components.} 
    \textbf{Green}: TP (Intersection), \textbf{Orange}: FP (Hallucination), \textbf{Red}: FN (Omission). 
    Visualizing the error components reveals that M2SVid yields a higher error driven by significant \textit{omission} of details (Red). We note that our model produces slightly more \textit{hallucinations} (Orange) than M2SVid. We are visualizing only every 10th match for readability purposes. The figure is best viewed zoomed in.}
    \label{fig:match_viz}
\end{figure*}

\input{src/tabs/headset_study}

\subsection{Matchability Interpretation and Visualization}
\label{supp:matchability}

In this section, we provide more intuitive understanding of the Matchability Error, and what it measures. 

Recall that the Matchability Error ($\mathcal{E}_{\text{Match}}$) is defined as the complement of the Jaccard index between the set of epipolarly-consistent matches in the ground truth ($M_{gt}$) and the prediction ($M_{pred}$):
\begin{equation}
    \mathcal{E}_{\text{Match}} = 1 - \frac{|M_{gt} \cap M_{pred}|}{|M_{gt} \cup M_{pred}|}.
\end{equation}
To analyze the specific nature of stereoscopic artifacts, we decompose this metric using True Positives ($N_{TP} = |M_{gt} \cap M_{pred}|$), False Positives ($N_{FP} = |M_{pred} \setminus M_{gt}|$), and False Negatives ($N_{FN} = |M_{gt} \setminus M_{pred}|$). The error can be expressed as:
\begin{equation}
    \mathcal{E}_{\text{Match}} = \frac{N_{FP} + N_{FN}}{N_{TP} + N_{FP} + N_{FN}}.
\end{equation}
This decomposition allows us to categorize the fidelity of the synthesis into three distinct components:

\begin{enumerate}
    \item \textbf{Omission Errors (False Negatives):} A high $N_{FN}$ indicates that the model fails to generate sufficient texture detail to support feature matching, often resulting from over-smoothing or blurring in complex regions.
    \item \textbf{Hallucination Errors (False Positives):} A high $N_{FP}$ indicates that the model generates high-frequency details that satisfy epipolar constraints but does not correspond to the ground truth scene structure (e.g., texture artifacts).
    \item \textbf{Feature Density (True Positives):} Crucially, the error rate must be contextualized by the absolute number of correct matches ($N_{TP}$). A method could artificially lower its error score by generating featureless (blurry) content, thereby minimizing both TP and FP. 
\end{enumerate}
A higher feature density ($N_{TP}$) combined with a low overall error indicates the successful preservation of sharp, matchable, and geometrically correct details.

\paragraph{Qualitative Analysis.} We illustrate the functioning of this metric with a qualitative example in Figure~\ref{fig:match_viz}. We visualize the ground truth left/right, as well as the right views generated by M2SVid or our approach for a particular example. From the zoomed-in example, it is evident that M2SVid loses a lot of the high-frequency details, making the text unreadable. For the generated views, we also show the true positive, false positive and false negative matches. We observe that M2SVid yields a higher error ($\mathcal{E}_{\text{Match}}=52.0$) driven primarily by a high rate of False Negatives (Red). This confirms a loss of high-frequency detail required for robust feature matching. In contrast, our method achieves a lower error ($\mathcal{E}_{\text{Match}}=40.9$) with a substantially higher density of True Positives (Green). While our method exhibits a slightly higher False Positive rate—indicating the generation of plausible but non-ground-truth textures—it successfully recovers significantly more valid structural details, resulting in superior overall stereoscopic fidelity.

\section{More Evaluation Results}
\label{sec:results-sup}

To verify that the quantitative superiority of our approach translates to better user experience on headsets, we performed a user study in Sec.~\ref{sec:user_study}. We also provide results according to the iSQoE metric in Sec.~\ref{sec:isqoe_results}. In Sec.~\ref{subsec:ego4d}, we present state-of-the-art results on the Ego4D dataset~\cite{ego4d}. Finally, in Sec.~\ref{sec:latency}, we compare the latency of our approach to alternative state-of-the-art methods.

\subsection{User Study on Headset}
\label{sec:user_study}

To evaluate the perceptual quality and geometric consistency of our generated videos, we conducted an anonymized blind pairwise comparison study on a headset. We recruited 9 participants to evaluate a diverse set of 26 scenes, selected from both the iPhone and Apple Vision Pro (AVP) datasets to ensure coverage of varying scene complexities.

\paragraph{Methodology.}
In each trial, participants were first presented with the Ground Truth (GT) video to establish a reference for correct geometry and visual details. Subsequently, they were shown two generated videos (3D visualization of the side-by-side): one produced by our method and one by a baseline method (either M2SVid or Eye2Eye). The position of the generated videos (Sample A vs. Sample B) was randomized to prevent bias. Participants were asked to select the video that exhibited better visual quality and 3D consistency compared to the GT reference, or to select ``Equal / No Preference'' if the results were indistinguishable. In total, we collected 120 pairwise judgments for each baseline comparison.

\paragraph{Results.}
The results, summarized in Table~\ref{tab:pairwise_study_results}, demonstrate a clear preference for our approach over both baselines.

When compared against M2SVid, our method achieved a dominant preference rate of 81.7\% (98 votes), while the baseline was preferred in only 1.7\% of cases.

Against Eye2Eye, the comparison was more competitive, reflecting Eye2Eye's high per-frame visual quality. However, our method was still preferred 50.0\% of the time -- four times more frequently than Eye2Eye (12.5\%). A significant portion of trials (37.5\%) resulted in a tie, suggesting that while Eye2Eye produces high-quality pixel-space outputs, our method achieves comparable or superior quality.

\subsection{Results on Ego4D} \label{subsec:ego4d}

We present results on Ego4D~\cite{ego4d} in Tab.~\ref{tab:ego4d_results}. Our approach obtains the best performance on all metrics for all categories. SVG, which relies on a pre-trained diffusion model without task-specific fine-tuning, demonstrates the poorest performance, ranking lowest in all categories. ReStereo, which combines video generation with restoration, often hallucinates textures, resulting in poor stereoscopic fidelity. Similarly, the latent diffusion-based StereoCrafter struggles with detail preservation due to lossy VAE encoding and often produces blurry outputs, resulting in poor P-PSNR scores and high Matchability Errors. 

Although M2SVid attempts to mitigate this via image-based losses, obtaining largely better P-PSNR and Matchability Error, it fails to fully overcome the VAE compression artifacts. In contrast, despite utilizing the same latent diffusion backbone, our approach generates sharp images with high-frequency details—evidenced by superior Matchability Error and P-PSNR results—thanks to our proposed guided decoder.

Furthermore, our direct generation approach achieves lower disparity errors compared to warping-based methods, which are inherently constrained by the limitations of off-the-shelf monocular depth estimators. 

Finally, by leveraging the proposed disparity conditioning, our approach ensures pixel-aligned outputs, leading to the best overall quality metrics.

\input{src/tabs/2c_main_ego4d}

\subsection{Results of the iSQoE Metric}
\label{sec:isqoe_results}

To further validate the perceptual quality of our generated stereo pairs, we utilize the iSQoE (Immersive Stereoscopic Quality of Experience) metric~\cite{isqoe}. Unlike traditional metrics that rely on pixel-wise differences, iSQoE is a no-reference, learning-based model explicitly trained to predict human preferences in Virtual Reality environments. It outputs a single scalar value where a lower score indicates higher quality.
In Tab.~\ref{tab:stereo4d_isqoe_results}, we report the iSQoE scores across the AVP, Stereo4D, and iPhone test sets.

\textbf{Metric Saturation Analysis.} The quantitative results reveal a critical insight regarding the metric's sensitivity in high-fidelity regimes. As shown in the table, the scores for all state-of-the-art methods are tightly clustered between $0.505$ and $0.521$. Most notably, the Ground Truth (GT) scores are numerically nearly indistinguishable from the generated outputs. For the iPhone dataset, M2SVid ($0.505$) and our model ($0.506$) both achieve lower (better) scores than the Ground Truth ($0.510$), which should pose a theoretical lower bound. On Stereo4D, our model and M2SVid tie for the best score ($0.515$).

This suggests that iSQoE heavily saturates when evaluating high-quality imagery. The metric was trained on the SCOPE dataset, which includes both traditional signal corruptions (noise, blur) and generative artifacts from methods like SDEdit and MotionCtrl. However, the metric appears to be highly tuned to the specific artifacts present in its training set and are presumably insensitive to inconsistencies of other latent video diffusion models. Effectively, once the image quality passes a certain threshold of ``cleanliness" -- a threshold met by all modern methods—the metric bottoms out around $0.50$, treating high-quality synthesis and real images as perceptually equivalent.

\input{src/tabs/1_stereo4d_isqoe}

\subsection{Validation of Disparity Control}
\label{sec:delta_validation}

A key property of our approach is the ability to control the strength of the generated stereo effect through the scalar conditioning signal $\delta$. To validate that this control is accurate and predictable, we performed 100 inference runs on 10 in-the-wild videos while varying $\delta$ across 10 values. We then estimated the resulting median disparity using FoundationStereo~\cite{foundationstereo}.

\begin{figure}[t]
  \centering
  \includegraphics[width=0.95\linewidth]{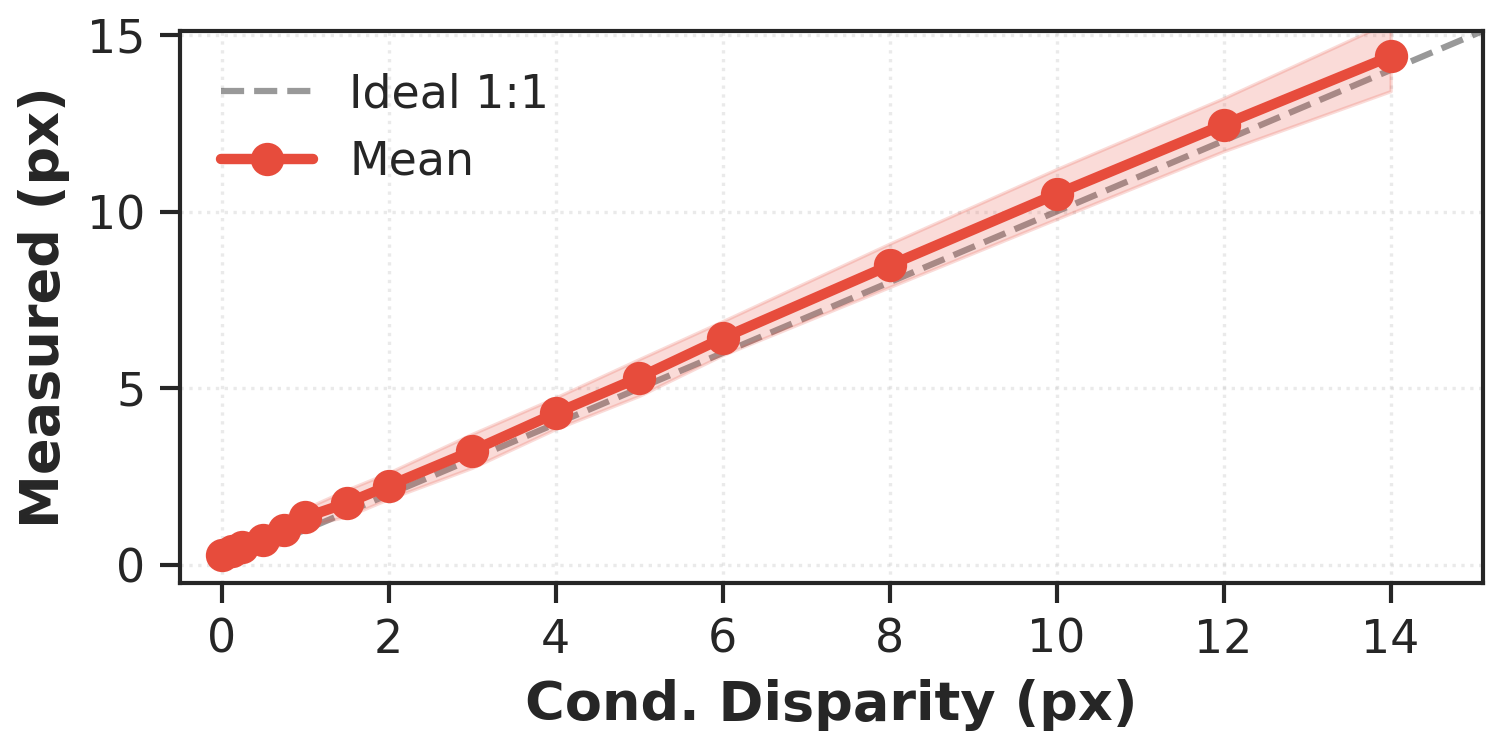}
  \caption{\textbf{Validation of disparity control.} 
  On the first frame, the generated median disparity closely matches the conditioning target, yielding an MAE of $0.54$ px. Across later frames, the generated stereo strength remains highly faithful to the conditioning signal, with an average linear correlation of $r = 0.986$ between target and measured median disparity.}
  \label{fig:sps_corr1}
\end{figure}

On the first frame, where the conditioning is defined, the generated median disparity matches the target closely, yielding a low mean absolute error (MAE) of $0.54$ pixels (Fig.~\ref{fig:sps_corr1}). For subsequent frames, absolute disparity values are less directly comparable because scene content and visible depth distributions evolve over time. We therefore evaluate whether the generated stereo strength remains \emph{faithful in a relative sense} by measuring the linear correlation between the conditioning value and the measured median disparity.

Averaged across all videos and 45 frame indices, we observe a near-perfect linear response with correlation coefficient $r = 0.986$. This confirms that our scalar conditioning provides stable and highly predictable control over the stereoscopic effect throughout the sequence, even though it is defined from the first frame only.

\subsection{Latency Comparison}\label{sec:latency}

In this section, we analyze the inference latency of our proposed method compared to state-of-the-art baselines. All evaluations were performed on a single NVIDIA H100 GPU, generating video sequences of 16 frames at a resolution of $512 \times 512$. Table~\ref{tab:latency} summarizes the computational cost breakdown.

\paragraph{Depth-Dependent Baselines (SVG, StereoCrafter, M2SVid).}
Existing multi-stage pipelines rely on explicit depth estimation and geometric warping as necessary preprocessing steps. Using DepthCrafter~\cite{depthcrafter} with 8 denoising steps, this introduces a fixed computational overhead of approximately $3.3$s for depth estimation and $0.9$s for warping. SVG incurs a significantly higher warping cost ($7.2$s) as it necessitates warping inputs 8 times. While M2SVid boasts a very fast base inference time ($0.8$s), its total runtime is dominated by these preprocessing bottlenecks, resulting in a total latency of $5.0$s.

\begin{figure*}[t!]
    \centering
    \includegraphics[width=\linewidth]{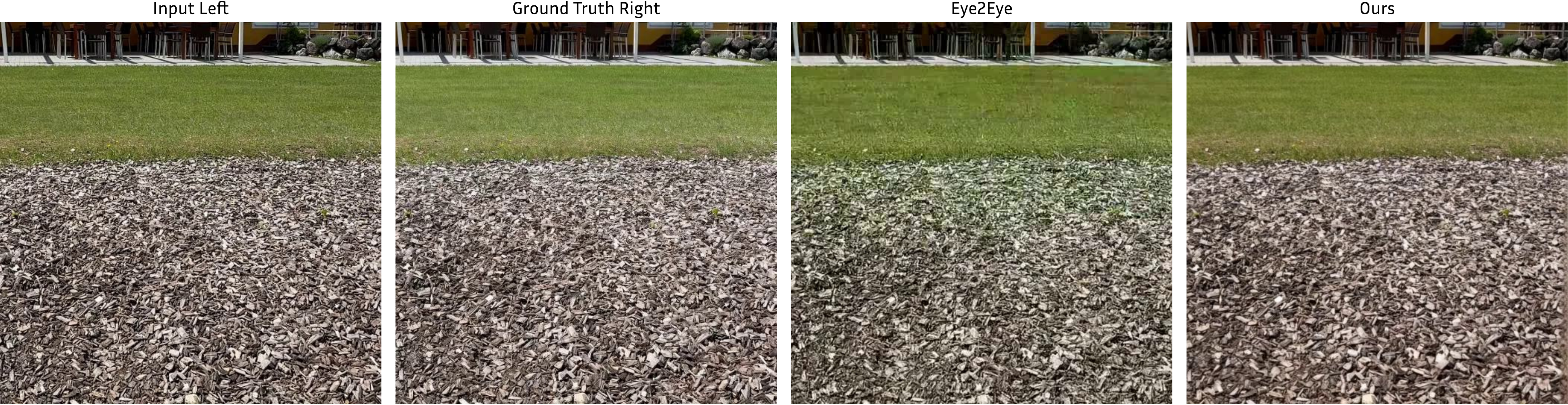}
    \caption{Eye2Eye exhibits color shifts.}
    \label{fig:eye2eye_colorshift}
\end{figure*}

\paragraph{Pixel-Space Diffusion (Eye2Eye).}
Comparison with Eye2Eye is challenging due to the non-public availability of the code; however, a theoretical analysis of computational complexity highlights significant efficiency gaps. First, the compute-intense Stage 2 of Eye2Eye operates in pixel space ($512 \times 512$) rather than the compressed latent space ($64 \times 64$) used by our method, resulting in a $64\times$ increase in spatial dimensionality. Second, it requires $50$ denoising steps compared to our single-step generation. A back-of-the-envelope calculation suggests a massive difference in computational load assuming similar model sizes:
\begin{equation}
    \underbrace{64\times}_{\text{Spatial Res.}} \times \underbrace{50\times}_{\text{Temporal Steps}} \approx 3200\times \text{ theoretical FLOPs increase.}
\end{equation}

\paragraph{Ours.}
It is observed that our model's pure inference time ($1.7$s) is higher than the M2SVid baseline ($0.8$s). The overhead comes from the added epipolar cross-attention and its associated feature extractor module.
In other words, we trade internal model complexity for pipeline simplicity, resulting in a net reduction in total latency.

\begin{table}[tb]
    \centering
    \caption{Model Latency Comparison. Runtimes are measured for generating a video of 16 frames with a resolution of $512 \times 512$. All evaluations were performed on an NVIDIA H100 GPU. ``Monodepth'' and ``Warp'' denote the time taken for depth estimation and geometric warping, respectively.}
    \label{tab:latency}
    \resizebox{0.45\textwidth}{!}{%
    \begin{tabular}{lcccc}
        \toprule
        \textbf{Method} & \textbf{Inference} & \textbf{Monodepth} & \textbf{Warp} & \textbf{Total} \\
        \midrule
        SVG             & 45s        & 3.3s      & 7.2s      & 55.5s \\
        StereoCrafter   & 2.6s       & 3.3s      & 0.9s      & 6.8s \\
        M2SVid          & 0.8s       & 3.3s      & 0.9s      & 5.0s \\
        \rowcolor{almond} Elastic3D  & 1.7s       & --        & --        & \textbf{1.7s} \\
        \bottomrule
    \end{tabular}
    }
\end{table}

\subsection{Analysis of Eye2Eye}

In this section, we provide additional analysis on competitor method Eye2Eye~\cite{eye2eye}. 

\parsection{Inference scheme}
Comparison with Eye2Eye required adaptations to match their fixed inference constraints.
First, the pre-trained Eye2Eye model accepts a fixed input length of 80 frames, whereas our evaluation benchmark consists of 16-frame clips. To accommodate this, we temporally padded the input by concatenating the clip and its reverse in a loop pattern $[V, V_{rev}, V, V_{rev}, V]$ to reach the required length. After inference, we extract the first 16 frames for evaluation.
Second, the trained Eye2Eye model is trained to generate a \textit{Left} view conditioned on a \textit{Right} view condition, which opposes our benchmark task (Left $\to$ Right). To bridge this gap, we horizontally flipped our input Left views before inference—effectively treating them as Right views—and subsequently flipped the generated outputs back to obtain the final Right view. We qualitatively verified that this mirroring process did not introduce specific artifacts, such as inverted text or texture distortions.

\parsection{Visual quality} We noticed that Eye2Eye generates views with widely different visual quality, depending on the inputs. On many samples, it generates high-quality outputs, with preserved texture and details. However, we also noticed that on a non-negligible amount of samples, it exhibits color shifts, as shown in Fig.~\ref{fig:eye2eye_colorshift}, presumably resulting from bleeding artifacts from neighboring frames. Moreover, on other examples, it produced blurry outputs, such as on the face of the man in the second row of Fig.~5 of the main paper, the swan of Fig.~\ref{fig:avp_121}, the leaves of Fig.~\ref{fig:iphone_56} or the text on the bus in Fig.~\ref{fig:iphone_55}.
These types of artifacts explain the relatively poor stereoscopic metrics that Eye2Eye obtains on average, in particular in terms of P-PSNR.

\parsection{Geometric correctness and 3D effect control} Eye2Eye does not have any mechanism to control the amount of 3D effect that it can generate. As a result, as evidenced by the different samples in Fig.~\ref{fig:avp_158}, ~\ref{fig:avp_29}, ~\ref{fig:avp_47} and ~\ref{fig:iphone_52}, it generates disparity maps in a different distribution than that of the ground-truth. Since the generated right views are not aligned to the ground truth, Eye2Eye obtains poor overall quality metrics (PSNR, SSIM, LPIPS), since those require pixel-wise alignment.

\section{More Ablation Studies}\label{sec:ablation-sup}

In this section, we provide ablation studies similar to those in the main paper on different datasets. We also ablate the signal used for conditioning.

\subsection{Warping-free conditioning} In Tab.~\ref{tab:ablation_conditioning_all}, we evaluate the impact of the proposed disparity conditioning (Sec.~4.2 of the main paper), which enables controlling the 3D strength in the generated stereoscopic videos, on the AVP, Stereo4D and iPhone datasets. Note that the results on Stereo4D and the iPhone datasets are already provided in Tab.~1 of the main paper and reported here for completeness. We train a baseline model without the disparity conditioning on the Stereo4D and Ego4D datasets, both of which contain videos predominantly captured with stereo cameras with baseline close to 63mm.
The results confirm that adding the conditioning results in better overall metrics for all datasets -- the gain is especially pronounced for the out-of-domain cameras setups.

\input{src/tabs/ablation_conditioning_all}

\begin{figure}[t]
    \centering
    \includegraphics[width=\linewidth]{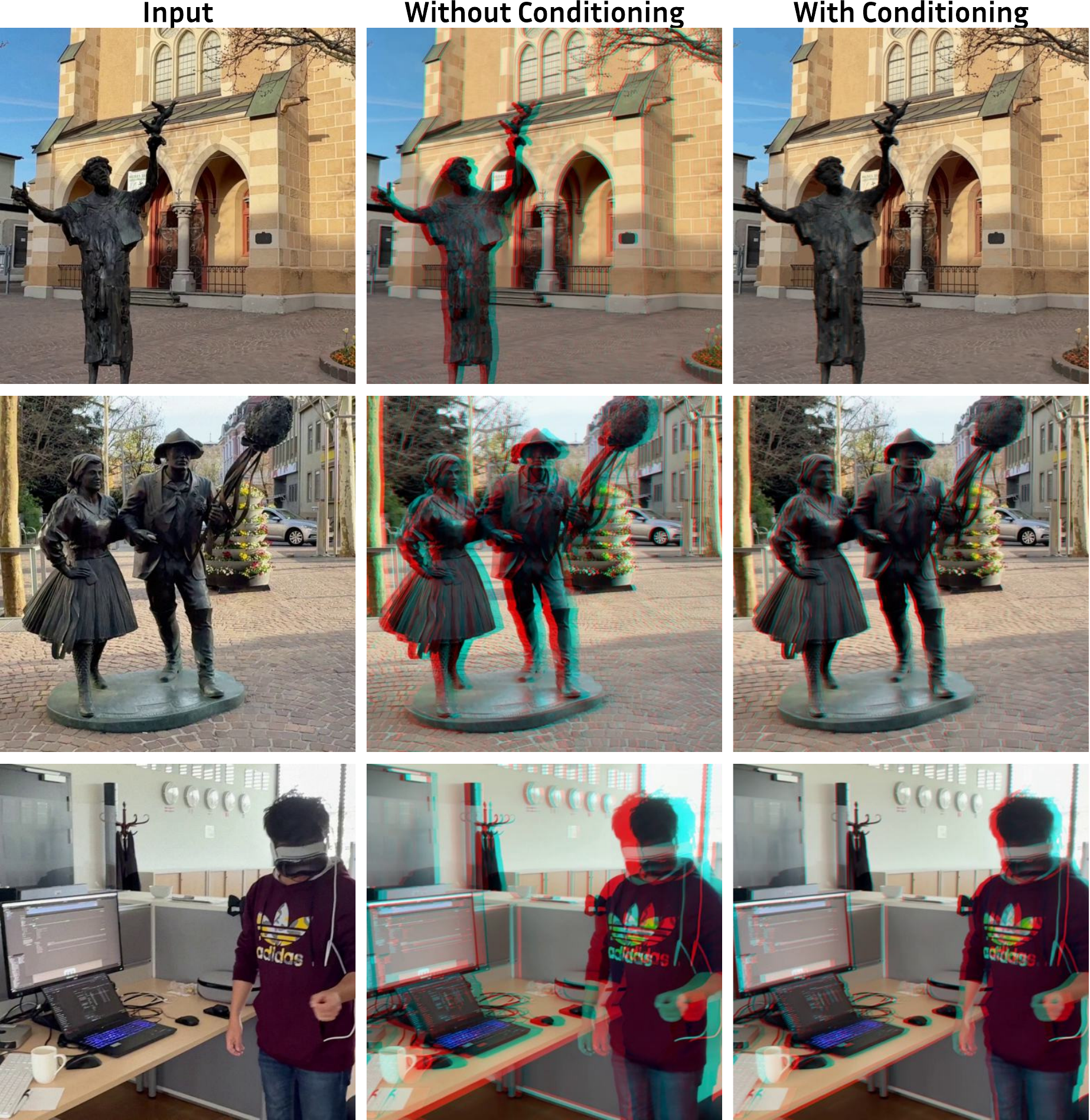}
    \caption{Input frames, and anaglyphs generated using our model trained without or with our proposed conditioning on the iPhone dataset. The iPhone's baseline of 19.2mm is significantly smaller than the one from the training datasets which is presumably ~63mm. It is evident that our conditioning approach is crucial for rescaling to desired baselines. }
    \label{fig:iphone_with_and_without_conditioning}
\end{figure}

In Fig.~\ref{fig:iphone_with_and_without_conditioning}, we qualitatively show the impact of the proposed disparity conditioning. We show the anaglyphs generated from the ground truth left and right view, our approach without or with conditioning. The baseline approach without conditioning has overfitted to the training baseline, close to ~63mm. It is thus incapable of generating content of the correct 3D strength on the iPhone dataset, which features a smaller baseline. Conversely, the proposed disparity conditioning allows the model to generate pixel-aligned right views for any disparity distributions.

\subsection{Conditioning types eq.~(3) of the main paper}

We compute the conditioning $\delta$ as the median of the disparity map relating the first frame of the left to the right video. In Tab.~\ref{tab:ablation_condition_type_all_revised}, we compare using the Median, Maximum or Average disparity as the conditioning signal, on the AVP, Stereo4D, iPhone and Ego4D datasets. All three types of conditioning obtain similar results on the four datasets. As a result, the median disparity was chosen as it is insensitive to outliers and provides an interpretable measure of the scene's overall stereo effect. However, both the average and the max disparity are also adequate choices.

\input{src/tabs/8_ablation_conditioning_types_all}

\subsection{Guided Latent Decoding} 

In this section, we evaluate the impact of the proposed guided decoder (Sec.~4.3 of the main paper) on the end stereo conversion task in Tab.~\ref{tab:ablation_study_model_vae_all}. We present results on the AVP, Stereo4D and iPhone datasets. Note that results on Stereo4D were already provided in Tab.~4 of the main paper and are added here for completeness. 

On all datasets, when integrated into our warping-free framework, the guided decoder significantly enhances image sharpness and high-frequency detail, resulting in a drastic reductions in LPIPS and increases of at least $+0.9 dB$ PSNR. It also creates substantial gains in stereoscopic fidelity: we observe improvements of at least $+1.2 dB$ in P-PSNR and massive reductions in Matchability Error, with a particularly impressive relative improvement of $58 \%$ on the iPhone dataset. This shows that our proposed guided decoder effectively minimizes binocular rivalry.

Furthermore, the proposed decoder serves as a generalized plug-and-play component. For instance, replacing the standard decoder in M2SVid~\cite{m2svid} with ours at inference time—without any retraining—improves performance across all metrics and datasets, yielding notably strong relative improvements in LPIPS and in Matchability Error. 

Finally, we note that the decoder maintains depth accuracy (Disp.~err) while slightly enhancing temporal stability, a benefit attributed to the consistent recovery of high-frequency details.

\input{src/tabs/ablation_vae_all}

\section{Qualitative results}
\label{sec:quali-sup}

 \paragraph{Video Results (HTML).}  As stereoscopic video generation involves temporal dynamics that are difficult to convey through static frames, the attached HTML video viewer provides a more comprehensive comparison.

 \paragraph{Static Results.} 
In Fig.~\ref{fig:avp_158}, ~\ref{fig:avp_29}, ~\ref{fig:avp_42}, ~\ref{fig:avp_47},~\ref{fig:avp_121} and ~\ref{fig:avp_22}, we compare the generated right view by our approach to state-of-the-art SVG, StereoCrafter, Eye2Eye and M2SVid on multiple samples from the AVP datasets. We also show the ground truth right view for reference.
Fig.~\ref{fig:iphone_14}, ~\ref{fig:iphone_30}, ~\ref{fig:iphone_49}, ~\ref{fig:iphone_52},~\ref{fig:iphone_55} and ~\ref{fig:iphone_56} show similar comparisons on the iPhone dataset instead.

By skimming through the examples, it is evident that Eye2Eye struggles to generate a view that is pixel-aligned to the ground truth. The disparity in all cases is very different from the ground truth, and the distribution varies a lot depending on the input. In contrast, thanks to our proposed disparity conditioning, our approach can generate right view for any desired disparity distribution. The geometric accuracy of warping-based approaches SVG, StereoCrafter and M2SVid entirely depends on the quality of the underlying monocular depth model that they use to warp the left video. As a result, the disparity maps, although mostly accurate, can show variable quality depending on the input samples. 

In terms of visual quality, our approach recovers better texture, and high-frequency details than counterparts. This is particularly evident on texts such as in Fig.~\ref{fig:avp_158}, or on vegetation/leaves as seen in Fig.~\ref{fig:iphone_49}, ~\ref{fig:iphone_52} and ~\ref{fig:iphone_56}.

In Fig.~\ref{fig:inference_matrix},~\ref{fig:inference_matrix3} and~\ref{fig:inference_matrix2}, we show the anaglyphs generated by our approach on in-the-wild images with varying disparity conditioning. As the pixel-disparity conditioning increases, the strength of the 3D effect also increases, as evidenced by the anaglyphs.

\def\GridImgW{3221}      
\def\GridImgH{1112}      
\def\GridTop{63}         
\def\GridLeft{50}        
\def\GridColW{512}       
\def\GridGapX{19}        

\newcommand{\ShowQualitativeResult}[6]{%
    \begin{figure*}[p]
        \centering
        \sbox0{\includegraphics{#1}}%
        
        \begin{tikzpicture}
            \node[anchor=south west, inner sep=0] (image) at (0,0) {\includegraphics[width=\textwidth]{#1}};
            
            \begin{scope}[x={(image.south east)}, y={(image.north west)}]
                \foreach \col in {0,1,2,3,4,5} {
                    \pgfmathsetmacro{\BoxX}{ (\GridLeft + \col * (\GridColW + \GridGapX) + #2) / \GridImgW }
                    \pgfmathsetmacro{\BoxW}{ #4 / \GridImgW }
                    \pgfmathsetmacro{\BoxTop}{ 1 - ((\GridTop + #3) / \GridImgH) }
                    \pgfmathsetmacro{\BoxH}{ #4 / \GridImgH }
                    
                    \draw[red, line width=1pt] (\BoxX, \BoxTop) rectangle (\BoxX + \BoxW, \BoxTop - \BoxH);
                }
            \end{scope}
        \end{tikzpicture}
        
        \vspace{1mm}
        
        \newcommand{\LocalCrop}[1]{%
            \edef\ColOffset{\fpeval{\GridLeft + ##1 * (\GridColW + \GridGapX)}}%
            \edef\TrimL{\fpeval{\ColOffset + #2}}%
            \edef\TrimT{\fpeval{\GridTop + #3}}%
            \edef\TrimR{\fpeval{\GridImgW - (\TrimL + #4)}}%
            \edef\TrimB{\fpeval{\GridImgH - (\TrimT + #4)}}%
            \includegraphics[%
                trim={\fpeval{\TrimL/\GridImgW}\wd0} 
                     {\fpeval{\TrimB/\GridImgH}\ht0} 
                     {\fpeval{\TrimR/\GridImgW}\wd0} 
                     {\fpeval{\TrimT/\GridImgH}\ht0},
                clip,
                width=0.16\textwidth 
            ]{#1}%
        }%

        \centering
        \setlength{\tabcolsep}{1pt}
        \begin{tabular}{cccccc}
            \LocalCrop{0} & \LocalCrop{1} & \LocalCrop{2} & \LocalCrop{3} & \LocalCrop{4} & \LocalCrop{5} \\
        \end{tabular}

        \vspace{-3mm}
        \caption{#5 \textit{Note: This figure has been compressed for arXiv submission. Please refer to the project page for the uncompressed version.}}
        \label{#6}
    \end{figure*}
}

\input{src/figs/3_qualitative_avp_suppl}

\input{src/figs/3_qualitative_iphone_suppl}

\begin{figure*}
    \centering
    \includegraphics[width=1\linewidth]{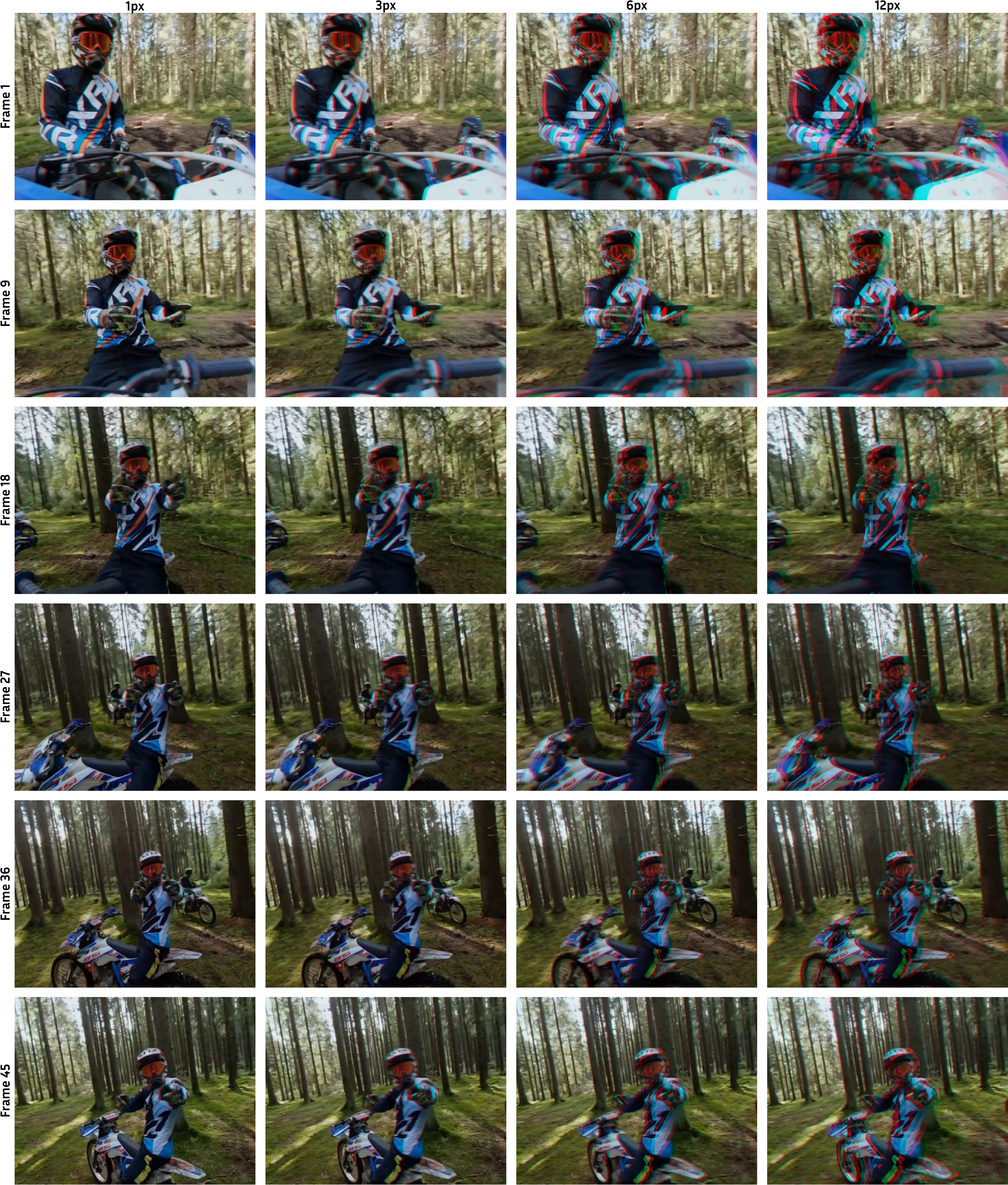}
    \vspace{-7mm}
    \caption{Anaglyphs generated by our approach on an in-the-wild image with varying disparity conditioning (in pixels). The results demonstrate the model's ability to control stereoscopic depth on real-world data. The frame number is indicated on the left while the disparity conditioning in pixel is written on the top row.}
    \label{fig:inference_matrix}
\end{figure*}

\begin{figure*}
    \centering
    \includegraphics[width=1\linewidth]{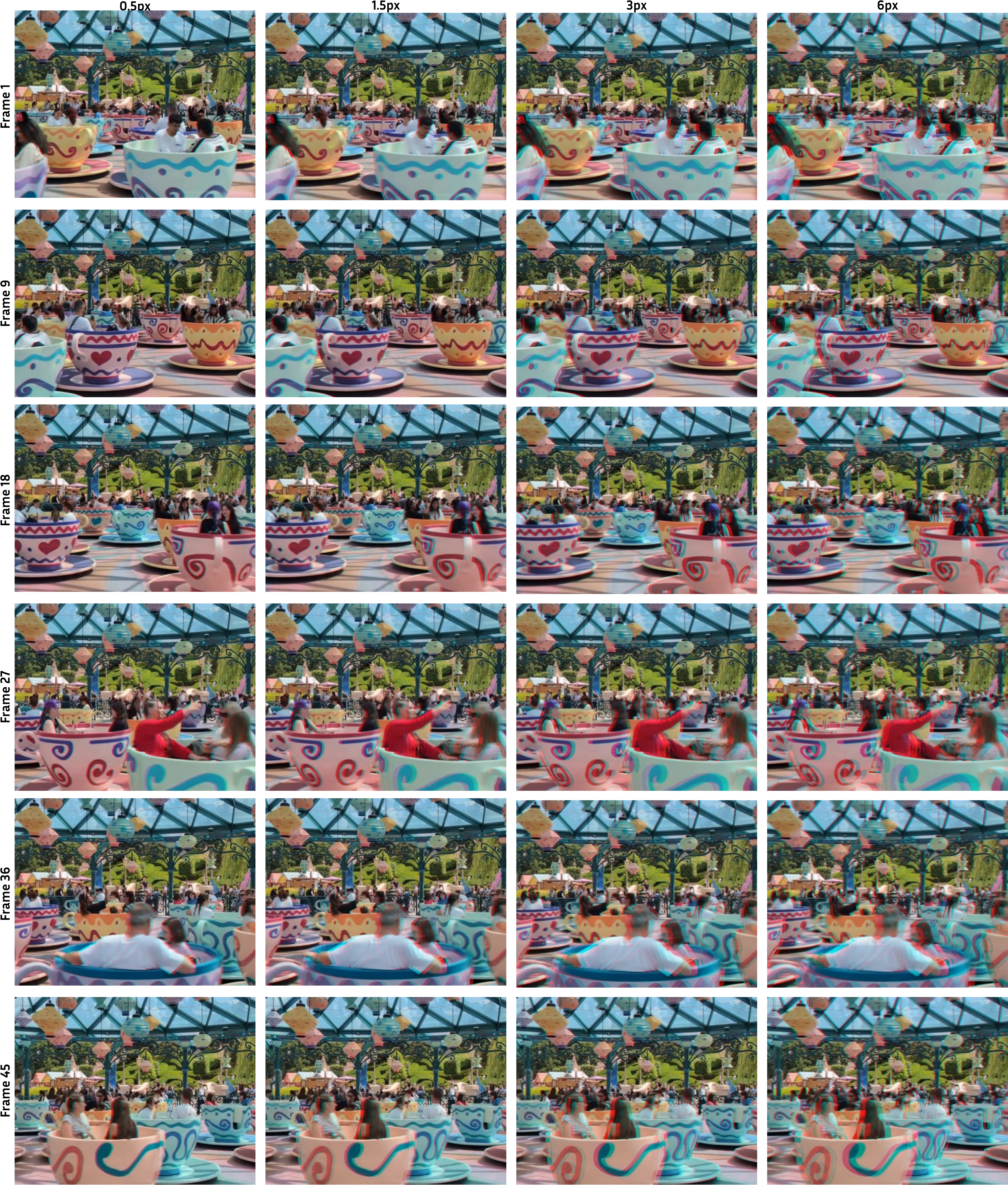}
    \vspace{-7mm}
    \caption{Anaglyphs generated by our approach on an in-the-wild image with varying disparity conditioning (in pixels). The results demonstrate the model's ability to control stereoscopic depth on real-world data. The frame number is indicated on the left while the disparity conditioning in pixel is written on the top row.}
    \label{fig:inference_matrix3}
\end{figure*}

\begin{figure*}
    \centering
    \includegraphics[width=1\linewidth]{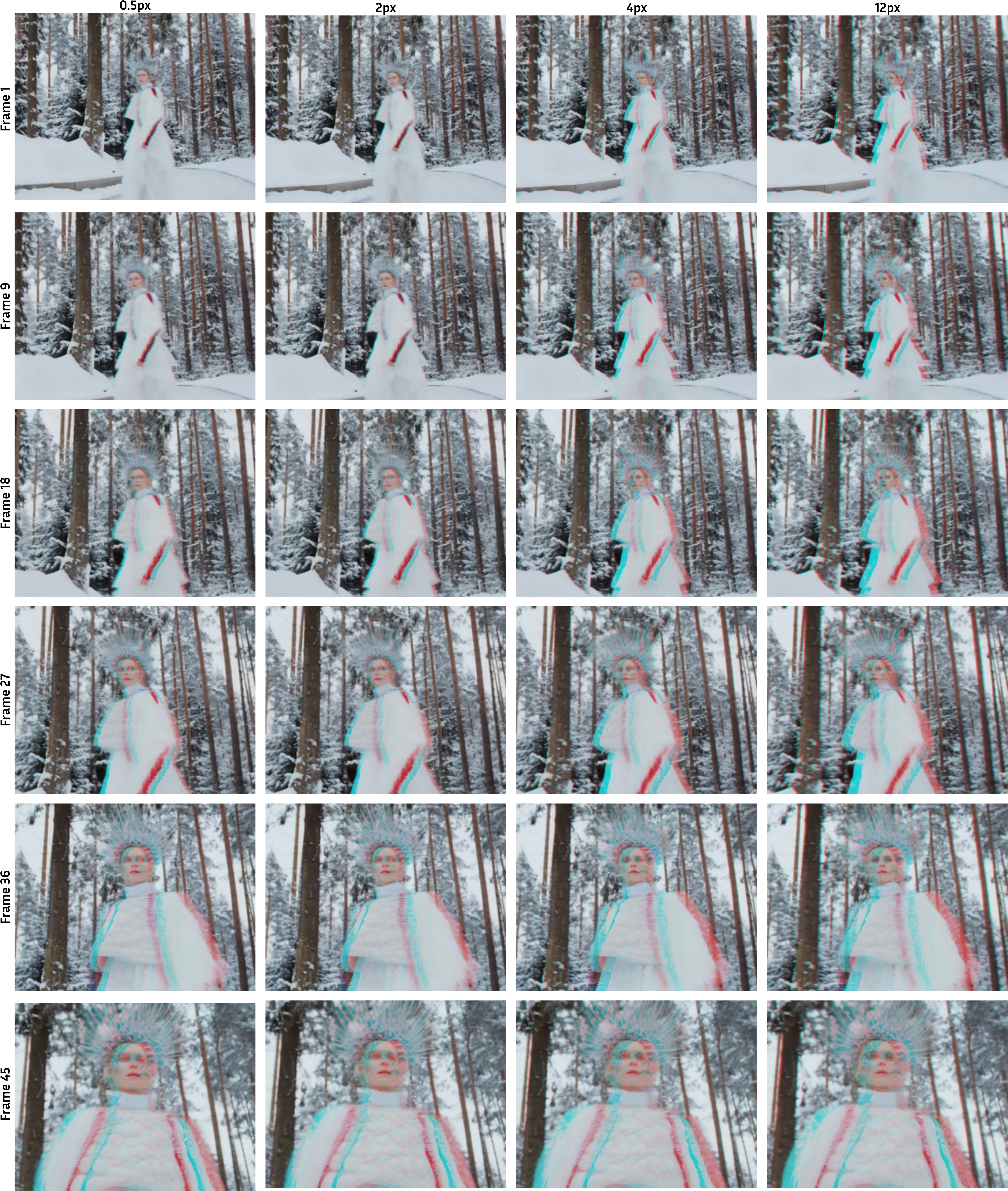}
    \vspace{-7mm}
    \caption{Anaglyphs generated by our approach on an in-the-wild image with varying disparity conditioning (in pixels). The results demonstrate the model's ability to control stereoscopic depth on real-world data. The frame number is indicated on the left while the disparity conditioning in pixel is written on the top row.}
    \label{fig:inference_matrix2}
\end{figure*}

\section{Limitations and Future Work}
\label{sec:future-work-sup}

While our method establishes a new baseline for warping-free stereoscopic video generation, several limitations remain which outline promising directions for future research.

\parsection{Inference on long and high-resolution videos}
Currently, our generation is constrained by GPU memory. On a single H100 (80GB) GPU, we can generate a maximum of approximately 45 frames at $512 \times 512$ resolution.
However, our methodology is theoretically extendable to arbitrary video lengths. Future work could adopt temporal autoregressive strategies similar to those employed by StereoCrafter~\cite{stereocrafter} and M2SVid~\cite{m2svid}, where the video is generated in chunks, conditioning the synthesis of the current chunk on the last $k$ frames of the previous one.
Similarly, higher-resolution inference could be achieved through spatial tiling and stitching strategies~\cite{stereocrafter,m2svid}, though we leave the implementation of these scaling techniques to future work.

\parsection{Extreme Disparities}
While our direct synthesis approach excels at standard stereoscopic baselines, it struggles to generalize to extreme disparities (baselines $\gg 63$mm). In these regimes, warp-based methods still hold an advantage, as they rely on explicit geometric reprojection rather than learned synthesis. Improving the data diversity to include wider baselines could mitigate this.

\parsection{Ambiguity of Median Conditioning}
Our use of a single scalar $\delta$ (median disparity) to control 3D strength is efficient but can be ambiguous. For example, a scene dominated by a flat background may have a median disparity near zero, even if foreground objects are present. As seen in Fig.~\ref{fig:inference_matrix2}, this ambiguity is pronounced in zoom-in shots: the sequence begins with a background-dominated view ($\delta \approx 0$), providing a weak conditioning signal. As the camera zooms into the subject and non-zero disparities appear, the model occasionally struggles to recover the correct stereo effect because the global conditioning token fails to capture the dynamic change in scene geometry.

\parsection{Geometric Inconsistencies}
While our warping-free approach eliminates the artifacts associated with occlusion filling and warping, it occasionally sacrifices strict geometric rigidity.
Warp-based methods rely on explicit, pre-trained depth estimators which enforce strong geometric priors — potentially ensuring, for instance, that flat walls remain perfectly planar.
In contrast, our model learns geometry implicitly from the training data.
As observed in our qualitative samples, this can sometimes lead to minor geometric hallucinations, such as ``wobbly'' depth on flat surfaces or inconsistent relative depths in complex scenes.
Since the mapping from a single image to a stereo pair is ill-posed, our model effectively predicts the most probable depth configuration, which may not always align perfectly with the physical ground truth.

\parsection{End-to-End Joint Training}
Currently, we train the U-Net and the Guided Decoder separately. While it is technically possible to train them jointly—allowing the U-Net to be optimized directly via the decoder's pixel-space reconstruction loss—we found this to be computationally prohibitive.
Joint training significantly increases the memory footprint, making the training-time versus memory trade-off impractical on standard hardware.
Furthermore, our factorized approach maintains modularity, allowing the Guided Decoder to be used as a plug-and-play enhancement for other latent diffusion pipelines.

\parsection{Computational Optimization}
As noted in our latency analysis, the current implementation of the epipolar cross-attention mechanism relies on a row-wise Python loop.
This introduces interpreter overhead and prevents full GPU saturation ($O(H)$ sequential iterations).
A significant speedup could be achieved by implementing this operation as a custom CUDA kernel, enabling fully parallelized execution.

\section{Acknowledgment}

Nina Shvetsova for the help throughout the project and for providing the initial codebase. Special thanks goes to Diego Martín Arroyo for helping with the infrastructure, experiments and technical discussion.
Thanks also to Haofei Xu, Felix Wimbauer, Luca Zanella, Nikolai Kalischek, Mattia Segu, Michael Niemeyer for technical inputs.

%% file: src/tabs/headset_study.tex
\begin{table*}[t!]
    \centering
    \caption{Results of Pairwise Human Perception Study. Participants compared \ours{} against two other methods: M2SVid and Eye2Eye. The table shows the number of times (\% of total for each row) each method was preferred, or if they were rated as equal.}
    \label{tab:pairwise_study_results}
    \vspace{-2mm}
    \resizebox{0.8\textwidth}{!}{ %
    \begin{tabular}{l c c c c}
        \toprule
        & \textbf{Total} & \multicolumn{3}{c}{\textbf{Preference in Pairwise Comparisons}} \\
        \cmidrule(lr){3-5}
        \textbf{Comparison (\ours vs.)} & \textit{(Count)} & \textbf{Competitor Preferred} & \textbf{Equal / No Preference} & \textbf{\ours{} Ours Preferred} \\
        & & \textit{(Count / \%)} & \textit{(Count / \%)} & \textit{(Count / \%)} \\
        \midrule
        M2SVid & 120 & 2 (1.7\%) & 20 (16.7\%) & \cellcolor{almond} \textbf{98} (\textbf{81.7\%}) \\
        Eye2Eye & 120 & 15 (12.5\%) & 45 (37.5\%) & \cellcolor{almond} \textbf{60} (\textbf{50.0\%}) \\
        \bottomrule
    \end{tabular}
    }
    \vspace{1mm}
\end{table*}

%% file: src/tabs/2c_main_ego4d.tex
\begin{table}[t]
  \centering
  \caption{Quantitative results on the Ego4D dataset.}
  \label{tab:ego4d_results}
  
  \resizebox{0.48\textwidth}{!}{%
  
  \begin{tabular}{@{}l@{}c@{~~~}c@{~~~}c|c@{~~~}c|c|c@{~}}
    \toprule
    \textbf{Method}  &  \textbf{PSNR} $\uparrow$ &  \textbf{SSIM} $\uparrow$ &  \textbf{LPIPS} $\downarrow$  &  \textbf{$\mathcal{E}_\text{Match}$} (\%) $\downarrow$ &  \textbf{P-PSNR} $\uparrow$  &  \textbf{MAE} $\downarrow$ &  \textbf{Flow  EPE} $\downarrow$  \\
    \midrule
    
    SVG                      & 12.7 & 0.294 & 0.595 & 71.7 & 14.6 & 11.9 & 49.8 \\
    StereoCrafter  & 16.1 & 0.539 & 0.450 & 50.5 & 17.4 & 8.10 & 8.10 \\
    M2SVid                & 18.0 & 0.681 & 0.334 & 43.6 & 24.6 & 5.28 & 6.59 \\
    ReStereo            & 15.2 & 0.477 & 0.479 & 44.9 & 17.3 & 11.2 & 8.53 \\
   
  \rowcolor{almond}   \ours                                & \textbf{19.8} & \textbf{0.760} & \textbf{0.276} & \textbf{27.0} & \textbf{25.9} & \textbf{3.30} & \textbf{5.67} \\

    \bottomrule
  \end{tabular}%
  }
\end{table}

%% file: src/tabs/1_stereo4d_isqoe.tex
\begin{table}[t]
    \centering
    \caption{\textbf{iSQoE Evaluation.} Comparison of stereoscopic quality on AVP, Stereo4D~\cite{stereo4d}, and iPhone datasets. \textbf{Lower is better} ($\downarrow$). The metric exhibits significant saturation near the optimal score.}
    \label{tab:stereo4d_isqoe_results}

    \resizebox{0.30\textwidth}{!}{%
    \begin{tabular}{@{}l c c c@{}}
        \toprule
        & \multicolumn{3}{c}{\textbf{iSQoE} $\downarrow$} \\
        \cmidrule(lr){2-4} %
        \textbf{Method} & \textbf{AVP} & \textbf{Stereo4D} & \textbf{iPhone} \\
        \midrule
        SVG             & 0.519          & 0.521        & 0.508  \\
        StereoCrafter   & \underline{0.509}          & 0.519        & 0.508  \\
        M2SVid          & \textbf{0.507}          & \textbf{0.515} & \textbf{0.505}  \\
        Eye2Eye         & \textbf{0.507} & \underline{0.517}        & 0.507  \\
        ReStereo        & --             & \underline{0.517}        & --      \\
        \cmidrule(lr){2-4}
        \rowcolor{almond} \ours & \underline{0.509}   & \textbf{0.515}        & \underline{0.506}  \\
        \midrule
        \textit{Ground Truth}       & \textit{0.505}   & \textit{0.513}        & \textit{0.510}  \\
        \bottomrule
    \end{tabular}%
    }
\end{table}

%% file: src/tabs/ablation_conditioning_all.tex
\begin{table}[t]
  \centering
  \caption{Impact of our conditioning approach (Sec.~4.2 of the main paper) on AVP Spatial Video (\textbf{top}), Stereo4D (\textbf{middle}) and iPhone Spatial Video (\textbf{bottom}).
  }\label{tab:ablation_conditioning_all}
  \vspace{-3mm}
  
  \resizebox{0.48\textwidth}{!}{%
  
  \begin{tabular}{@{}l@{}c@{~~~}c@{~~~}c@{~}|@{~}c@{~~~}c@{~}|@{~}c@{~}|@{~}c@{~}}
    \toprule
    \textbf{Method}  &  \textbf{PSNR} $\uparrow$ &  \textbf{SSIM} $\uparrow$ &  \textbf{LPIPS} $\downarrow$  &  \textbf{$\mathcal{E}_\text{Match}$} $\downarrow$ &  \textbf{P-PSNR} $\uparrow$  &  \textbf{Disp.\ err} $\downarrow$ &  \textbf{Temp.\ err} $\downarrow$  \\
    \midrule

    \ours w/o Cond          & 21.1 & 0.710 & 0.301 &  \textbf{23.4} & 27.6  & 3.11 & 1.70 \\
    
    \ours (ours)                    & \textbf{25.9} & \textbf{0.894} & \textbf{0.196} & 30.9 & \textbf{28.4} & \textbf{1.74} & \textbf{1.31} \\

    \midrule

    \ours w/o Cond          &  25.1 & 0.880 & 0.192 & 28.6 & 27.2 & 1.27 & \textbf{1.28} \\
    
    \ours (ours)            &  \textbf{26.1} & \textbf{0.913} & \textbf{0.176}  & \textbf{27.8} & \textbf{27.4} & \textbf{1.24} & 1.30 \\

    \midrule

    \ours w/o Cond         & 18.7 & 0.703 & 0.289 & 30.3 & 25.2 & \textbf{0.64} & 3.23 \\
    
    \ours (ours)            & \textbf{22.5} & \textbf{0.890} & \textbf{0.193} & \textbf{26.5} & \textbf{26.2} & 0.77 & \textbf{3.10} \\
    
    \bottomrule
  \end{tabular}%
  }

\end{table}

%% file: src/tabs/8_ablation_conditioning_types_all.tex
\begin{table}[t]
  \centering
  \caption{Impact of the disparity conditioning types (Sec.~4.2 of the main paper) on the AVP  Spatial Video, Stereo4D, iPhone Spatial Video and Ego4D datasets.}
  \label{tab:ablation_condition_type_all_revised}
  
  \resizebox{0.41\textwidth}{!}{%
    \begin{tabular}{llccc}
      \toprule
      \textbf{Dataset} & \textbf{Model} & \textbf{PSNR} $\uparrow$ & \textbf{SSIM} $\uparrow$ & \textbf{LPIPS} $\downarrow$ \\
      \midrule
      
      \multirow{3}{*}{{AVP}} 
      & Average & \textbf{26.2} & \textbf{0.903} & \textbf{0.194} \\
      & Max     & 25.2         & 0.860         & 0.217 \\
      & Median  & 25.9         & 0.894         & 0.196 \\
      \cmidrule(lr){2-5} %
      
      \multirow{3}{*}{{Stereo4D}} 
      & Average & \textbf{26.4} & \textbf{0.913} & \textbf{0.176} \\
      & Max     & 25.6         & 0.897         & 0.185 \\
      & Median  & 26.1         & \textbf{0.913} & 0.\textbf{176} \\
      \cmidrule(lr){2-5} %
      
      \multirow{3}{*}{{iPhone}} 
      & Average & \textbf{23.4} & \textbf{0.916} & \textbf{0.180} \\
      & Max     & 22.9         & 0.902         & 0.184 \\
      & Median  & 22.5         & 0.890         & 0.193 \\
      \cmidrule(lr){2-5} %
      
      \multirow{3}{*}{{Ego4d}} 
      & Average & 19.6         & 0.751         & 0.290 \\
      & Max     & 19.6         & 0.754         & 0.293 \\
      & Median  & \textbf{19.8} & \textbf{0.760} & \textbf{0.276} \\
      
      \bottomrule
    \end{tabular}%
  }
\end{table}

%% file: src/tabs/ablation_vae_all.tex
\begin{table}[t]
  \centering
  \caption{Impact of our guided-VAE decoder $\mathcal{D'}$ (Sec.~4.3 of the main paper) for mono-to-stereo conversion on the AVP Spatial Video (\textbf{top}), Stereo4d (\textbf{middle}) and iPhone  Spatial Video (\textbf{bottom}) datasets.}
  \label{tab:ablation_study_model_vae_all}
  \vspace{-3mm}
  \resizebox{0.48\textwidth}{!}{%
  \begin{tabular}{@{}l@{}c@{~~~}c@{~~~}c@{~}|@{~}c@{~~~}c@{~}|@{~}c@{~}|@{~}c@{~}}
    \toprule
    
    \textbf{Model} & \textbf{PSNR} $\uparrow$ & \textbf{SSIM} $\uparrow$ & \textbf{LPIPS} $\downarrow$  &\textbf{$\mathcal{E}_\text{Match}$}  $\downarrow$& \textbf{P-PSNR} $\uparrow$ & \textbf{Disp. err} $\downarrow$ & \textbf{Temp. err} $\downarrow$\\
    \midrule
    \midrule

    \ours w/o $\mathcal{D'}$     & 25.5 & 0.884 & 0.229  & 44.9 & 27.3 & \textbf{1.71}  & 1.38 \\
    \ours (ours)     & \textbf{25.9} & \textbf{0.894} & \textbf{0.196} & \textbf{30.9} & \textbf{28.4} & 1.74 & \textbf{1.31} \\
    \hdashline
    
    M2SVid                        & 24.4 & 0.821 & 0.221 & 41.5 & 27.3 & \textbf{2.30} & 1.35 \\
    M2SVid + $\mathcal{D'}$       & \textbf{24.5} & \textbf{0.827} & \textbf{0.198} & \textbf{26.8} & \textbf{28.1 }& 2.33 & \textbf{1.32} \\

    \midrule
    \midrule
    
    \ours w/o $\mathcal{D'}$  & 25.2 & 0.895 & 0.212  & 41.9 & 26.1 & \textbf{1.23} & 1.37 \\

    \ours (ours) &  \textbf{26.1} & \textbf{0.913} & \textbf{0.176} & \textbf{27.8} & \textbf{27.4} & 1.24 & \textbf{1.30} \\
    
    \hdashline
    
    M2SVid                      &   24.6  &  0.819  &  0.206  &  39.6  &  26.3   &  \textbf{1.56}  &  1.35  \\
    M2SVid + $\mathcal{D'}$     & \textbf{25.2} & \textbf{0.832} & \textbf{0.175}  &  \textbf{24.8} & \textbf{27.5} & 1.61 & \textbf{1.27} \\

    \midrule
    \midrule
    
    \ours w/o $\mathcal{D'}$     & 21.9 & 0.868 & 0.239 & 62.5 & 24.7 & \textbf{0.74} & 3.34 \\
    \ours (ours)                        & 22.5 & \textbf{0.890} & \textbf{0.193} & \textbf{26.5} & \textbf{26.2} & 0.77 & \textbf{3.10}  \\
    \hdashline
    
    M2SVid                          & 22.9 & 0.865 & 0.205 & 38.4 & 25.1 & \textbf{0.60} & 3.06  \\ 
    M2SVid + $\mathcal{D'}$         & \textbf{23.9} & \textbf{0.892} & \textbf{0.158} & \textbf{21.0}  & \textbf{26.8} & \textbf{0.60} &\textbf{2.58} \\

    \bottomrule
    \bottomrule
  \end{tabular}%
  }
\end{table}

%% file: src/figs/3_qualitative_avp_suppl.tex
\ShowQualitativeResult
    {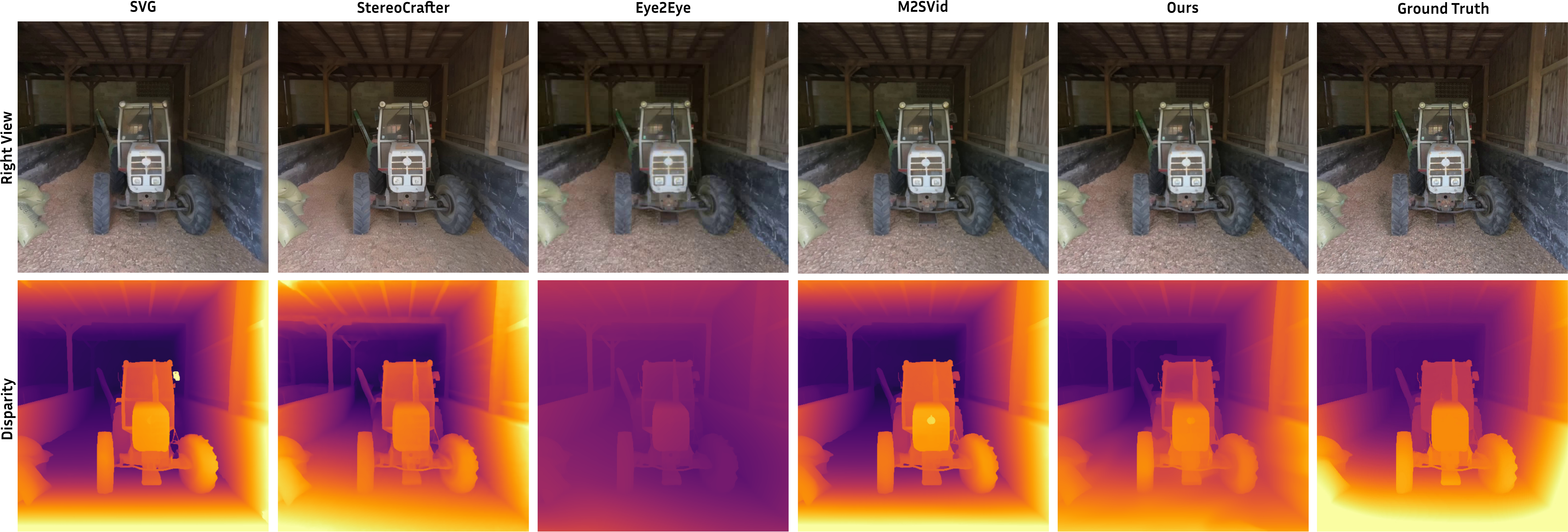} %
    {0} %
    {350} %
    {60}  %
    {\textbf{Comparison on Scene 158 of the AVP dataset.} Note the more readable text on the bag and the texture of the ground.} 
    {fig:avp_158}

\ShowQualitativeResult
    {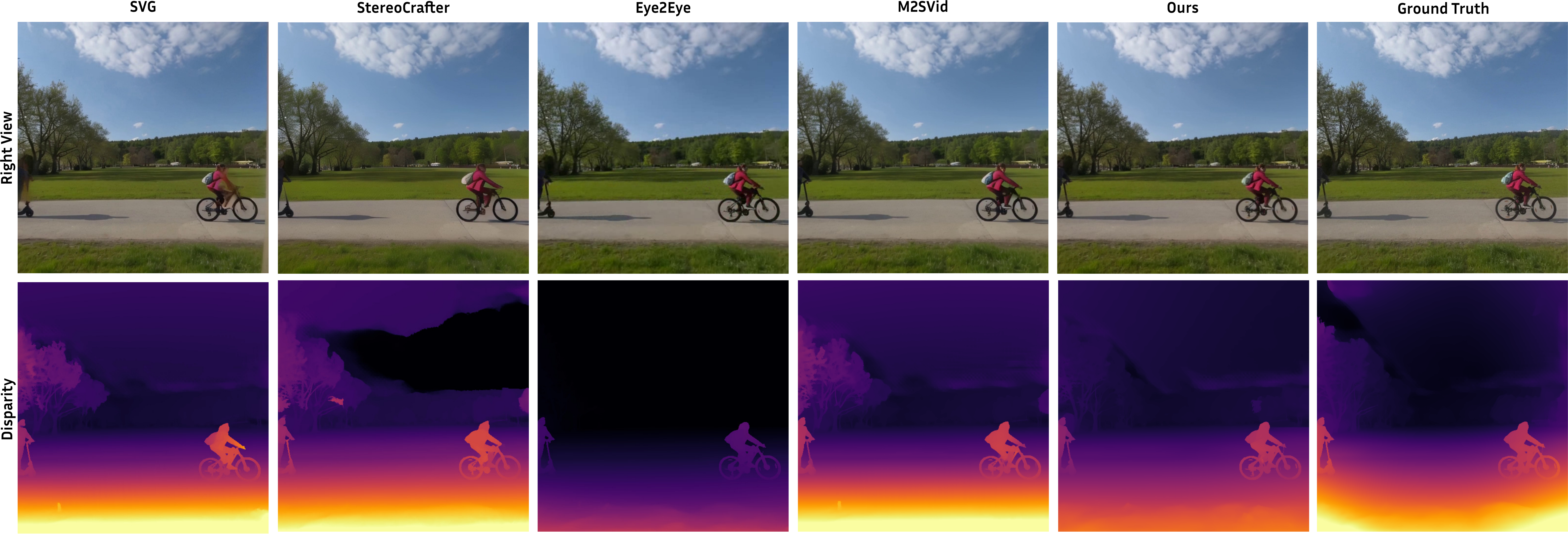} 
    {10}  %
    {180} %
    {70} %
    {\textbf{Comparison on Scene 29 of the AVP dataset.} Our method preserves the branch structures in the trees.} 
    {fig:avp_29}

\ShowQualitativeResult
    {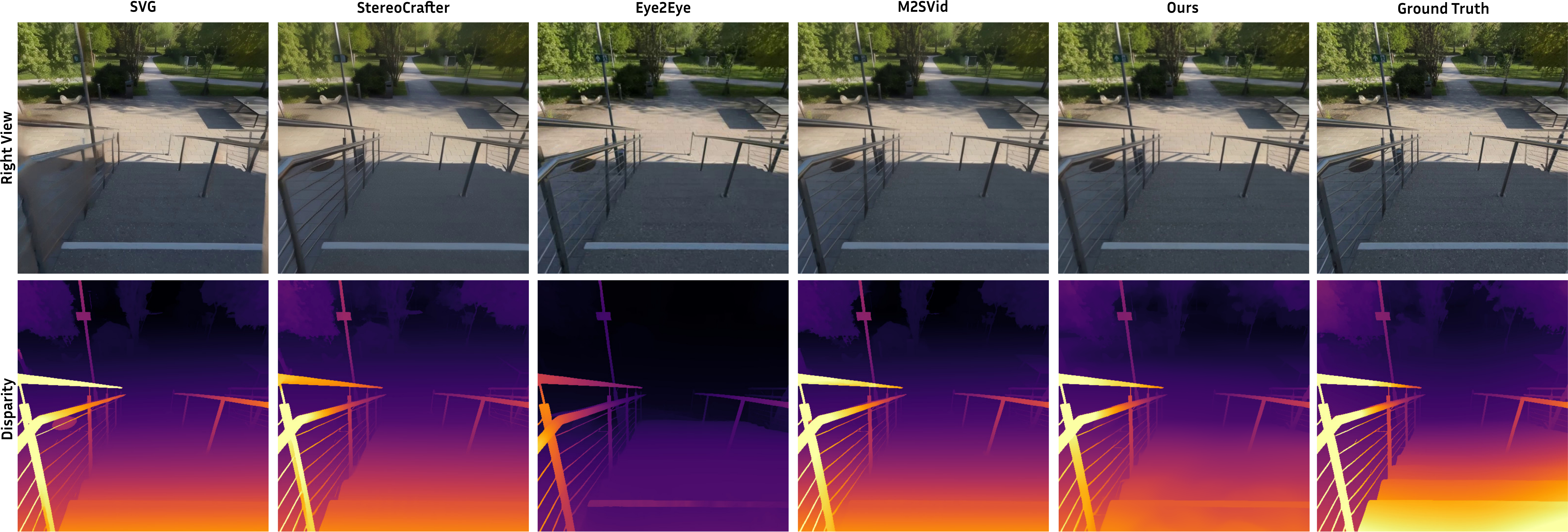} 
    {200}  %
    {50} %
    {75} %
    {\textbf{Comparison on Scene 42 of the AVP dataset.} Our method better preserves the tree and background.} 
    {fig:avp_42}

\ShowQualitativeResult
    {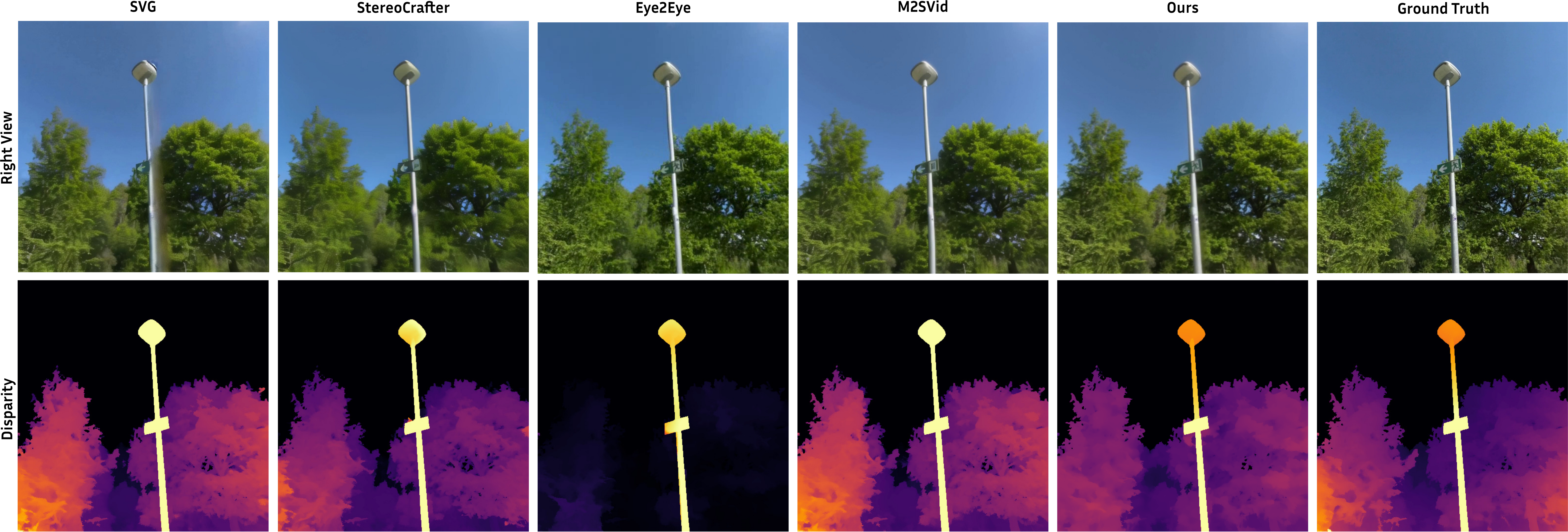} %
    {230} %
    {260} %
    {80}  %
    {\textbf{Comparison on Scene 47 of the AVP dataset.} Our method recovers the details on the sign.} 
    {fig:avp_47}

\ShowQualitativeResult
    {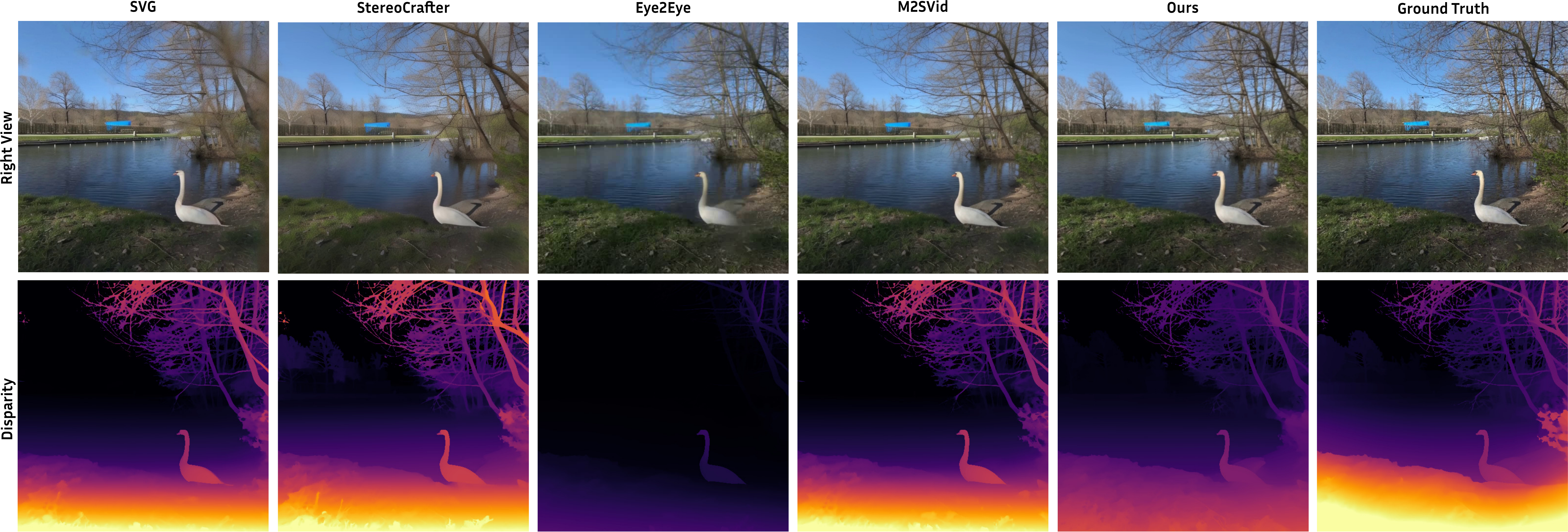} %
    {280} %
    {280} %
    {80}  %
    {\textbf{Comparison on Scene 121 of the AVP dataset.} Our method generates a sharp output, as evidenced by the swan head.} 
    {fig:avp_121}

\ShowQualitativeResult
    {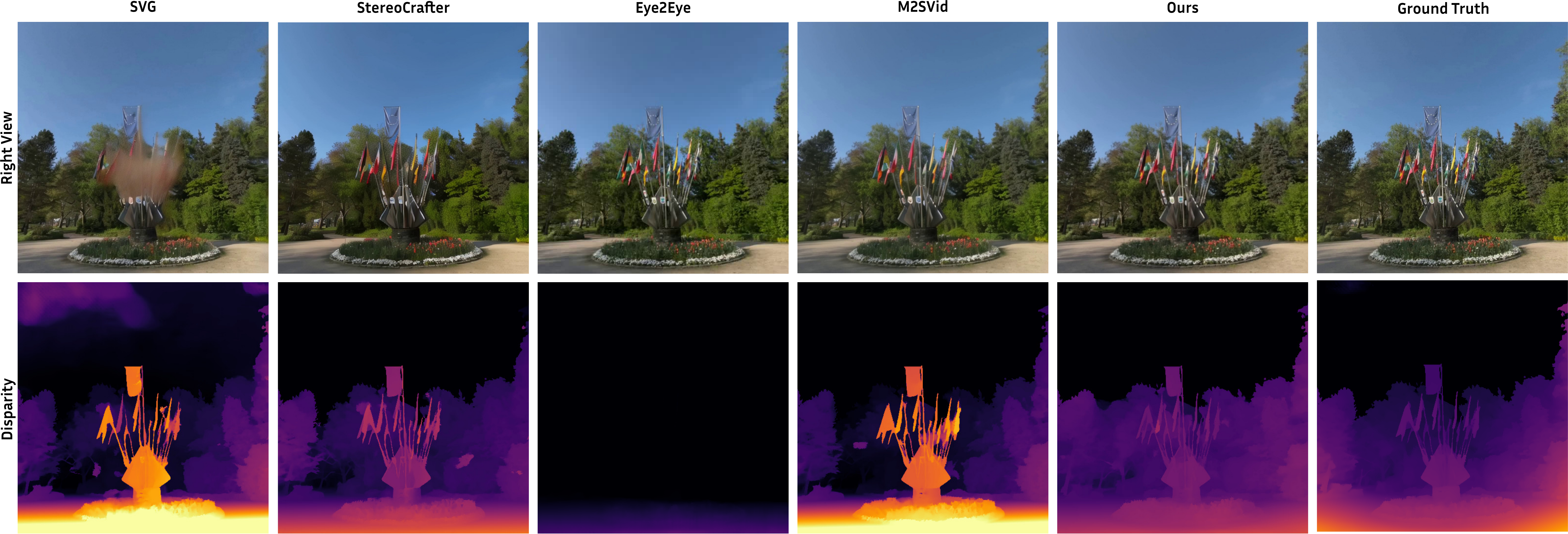} %
    {220} %
    {320} %
    {60}  %
    {\textbf{Comparison on Scene 22 of the AVP dataset.} Our method recovers the high-frequency details on the signs and better geometry.} 
    {fig:avp_22}

%% file: src/figs/3_qualitative_iphone_suppl.tex
\ShowQualitativeResult
    {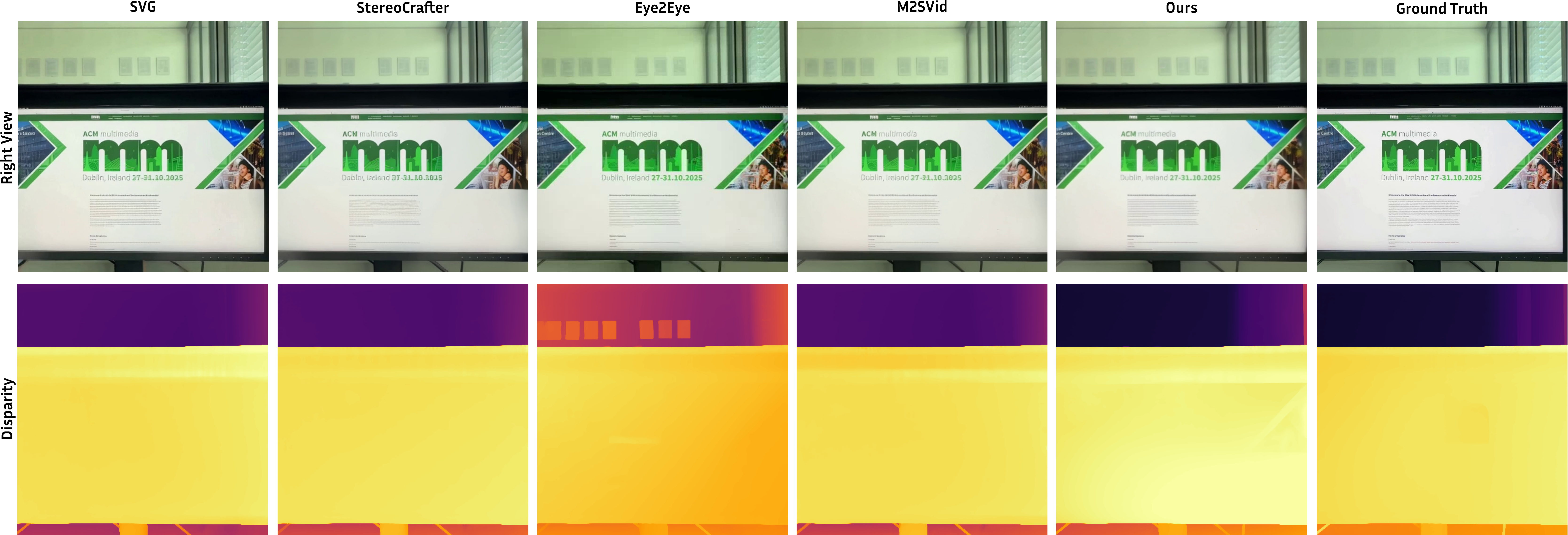} %
    {160} %
    {280} %
    {80}  %
    {\textbf{Comparison on Scene 14 of the iPhone dataset.} Our method produces readable text.} 
    {fig:iphone_14}

\ShowQualitativeResult
    {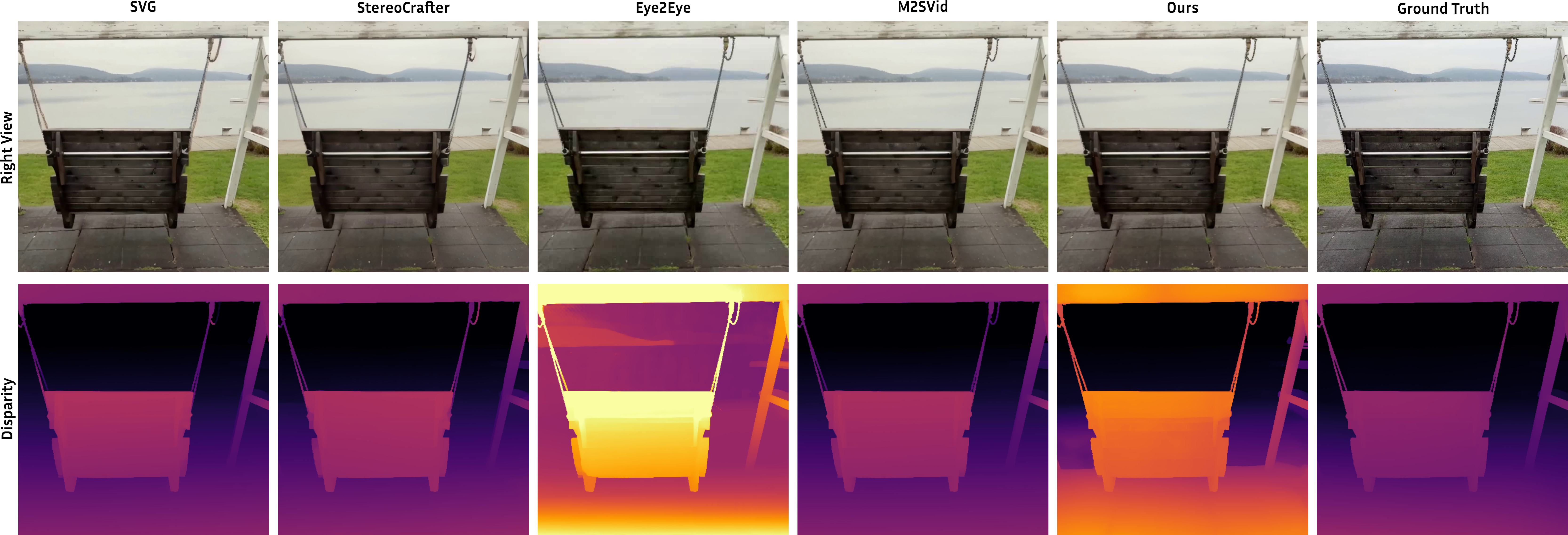} %
    {20} %
    {150} %
    {80}  %
    {\textbf{Comparison on Scene 30 of the iPhone dataset.} Our method recovers high-frequency details on the chain.} 
    {fig:iphone_30}

\ShowQualitativeResult
    {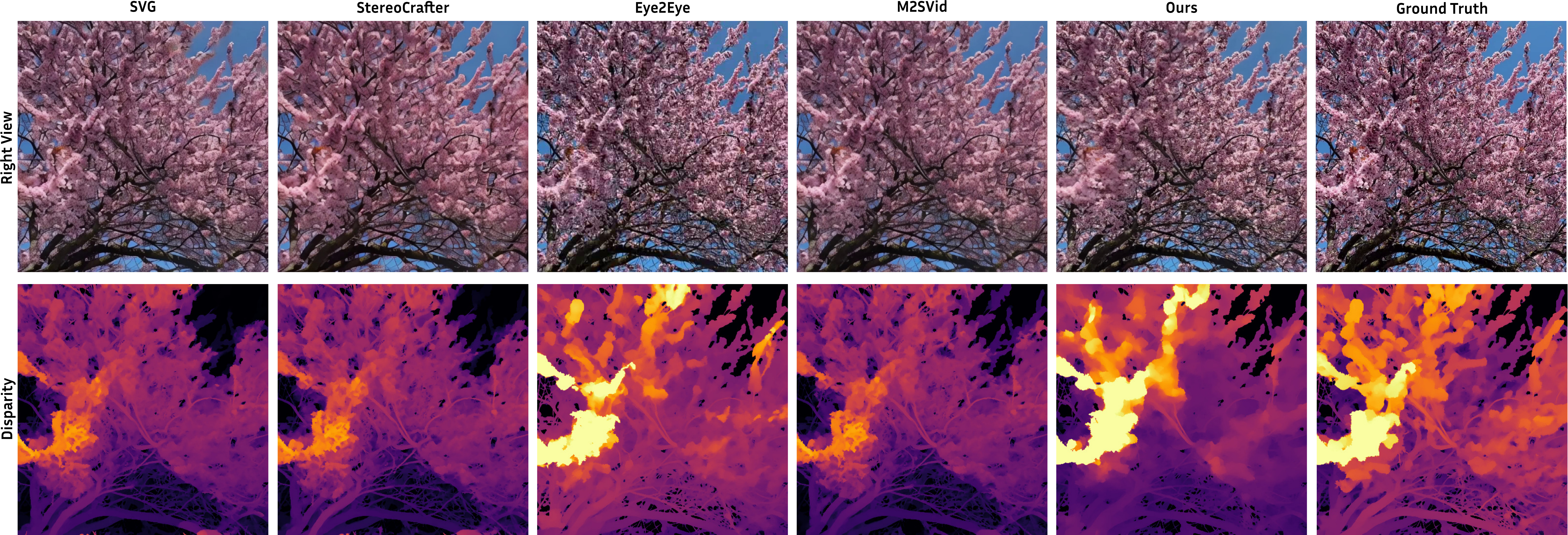} %
    {200} %
    {360} %
    {120}  %
    {\textbf{Comparison on Scene 49 of the iPhone dataset.} Our method recovers the texture on the leaves.} 
    {fig:iphone_49}

\ShowQualitativeResult
    {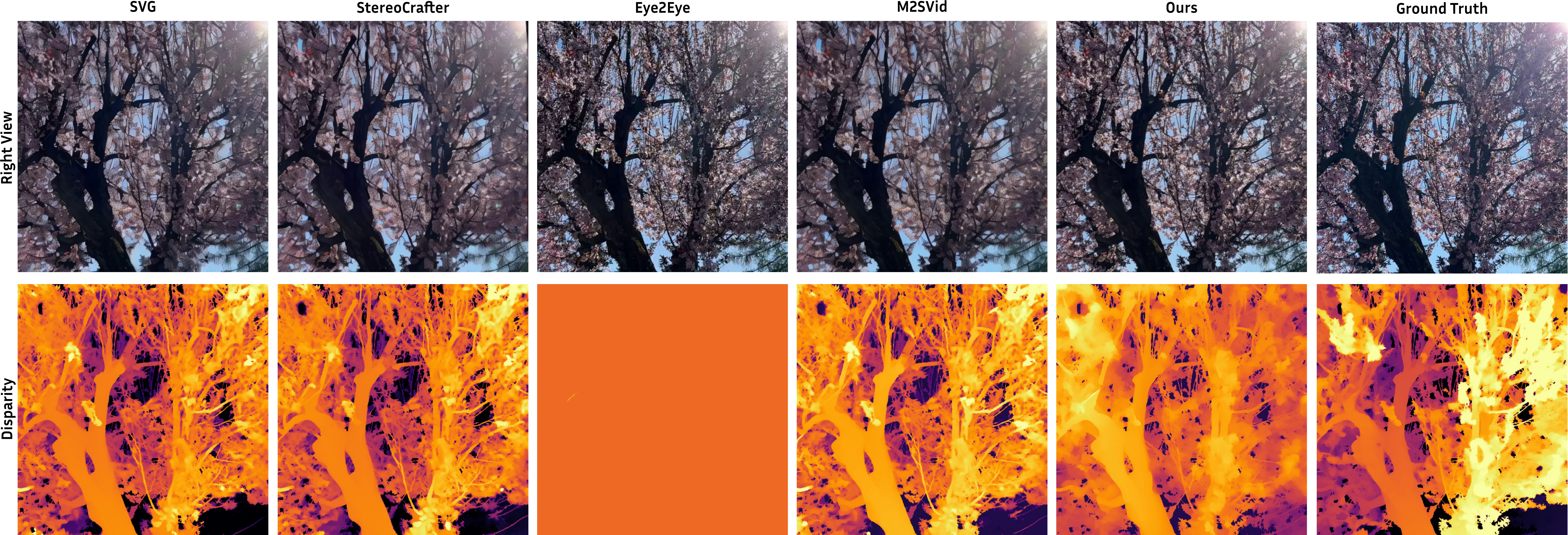} %
    {200} %
    {150} %
    {80}  %
    {\textbf{Comparison on Scene 52 of the iPhone dataset.} Our method recovers the texture on the leaves.} 
    {fig:iphone_52}

\ShowQualitativeResult
    {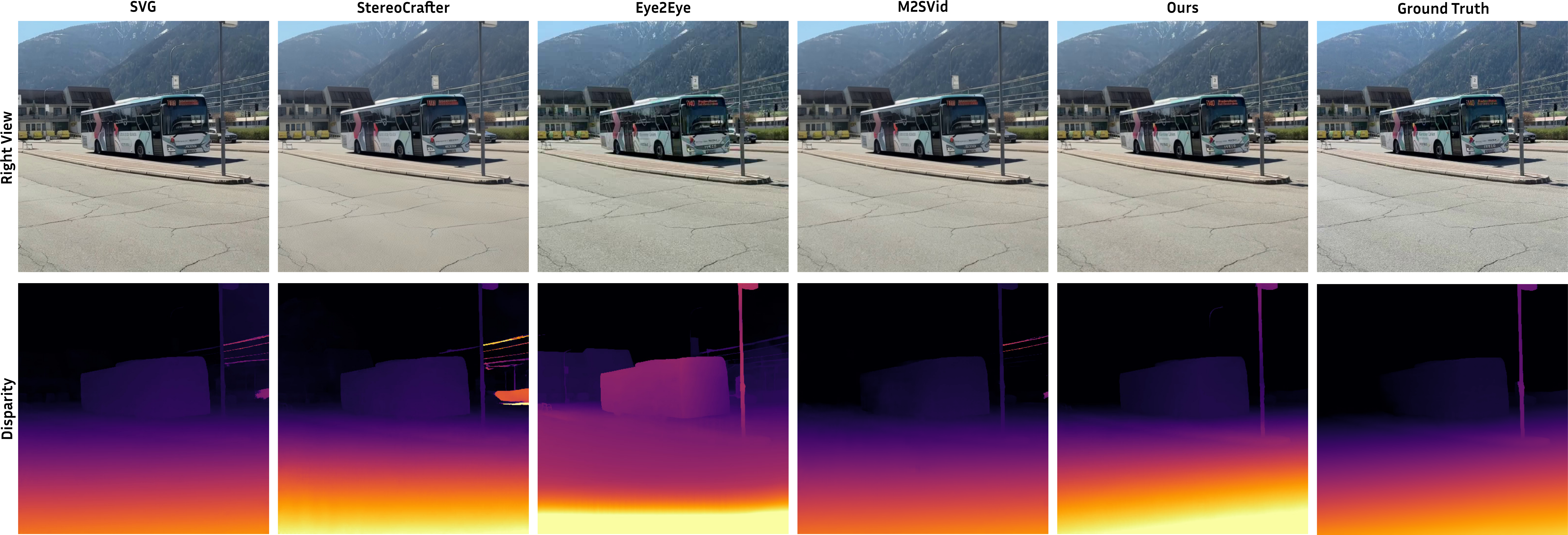} %
    {200} %
    {190} %
    {80}  %
    {\textbf{Comparison on Scene 55 of the iPhone dataset.} Our method preserves better the high-frequency details on text.} 
    {fig:iphone_55}

\ShowQualitativeResult
    {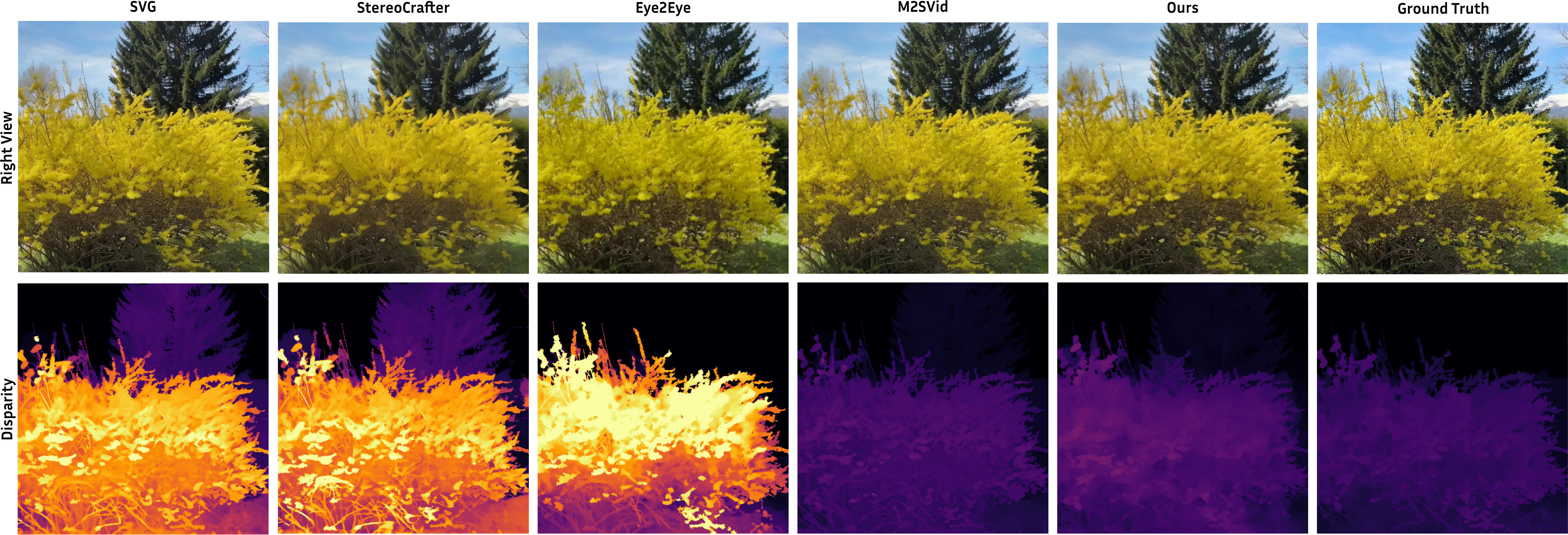} %
    {200} %
    {150} %
    {80}  %
    {\textbf{Comparison on Scene 56 of the iPhone dataset.} Our method produces sharper details and texture.} 
    {fig:iphone_56}